\documentclass{article}
\usepackage[utf8]{inputenc}
\usepackage{verbatim}
\usepackage[sort&compress,square,numbers]{natbib}
\usepackage{graphicx}
\usepackage{xcolor}
\usepackage[frozencache,cachedir=minted_cache]{minted}

\usepackage{hyperref}
\usepackage{sistyle}
\usepackage[symbol]{footmisc}

\SIthousandsep{,}
\hypersetup{
    colorlinks=true,
    linkcolor=blue,
    filecolor=magenta,      
    urlcolor=cyan,
    pdftitle={ELM},
    pdfpagemode=FullScreen,
}
\title{Evolution through Large Models}
\author{
  Joel Lehman\\
  OpenAI \\
  \texttt{joel@openai.com}
  \and
  Jonathan Gordon\\
  OpenAI \\
  \texttt{gordonjo@openai.com}
  \and
  Shawn Jain\\
  OpenAI\\
  \texttt{jains@openai.com}
  \and
  Kamal Ndousse\\
  Anthropic\footnote{Work done at OpenAI.}\\
  \texttt{kamal.ndousse@gmail.com}
  \and
  Cathy Yeh\\
  OpenAI\\
  \texttt{cathy@openai.com}
  \and
  Kenneth O. Stanley\\
  OpenAI\\
  \texttt{kennethostanley@gmail.com}
}
\date{\today}

\usepackage{amsmath,amsfonts,bm}

\def\eqref#1{equation~\ref{#1}}

\def\1{\bm{1}}

\def\vtheta{{\bm{\theta}}}

\def\vx{{\bm{x}}}

\DeclareMathAlphabet{\mathsfit}{\encodingdefault}{\sfdefault}{m}{sl}
\SetMathAlphabet{\mathsfit}{bold}{\encodingdefault}{\sfdefault}{bx}{n}

\def\sR{{\mathbb{R}}}

\usepackage{caption}
\usepackage{subcaption}
\usepackage[capitalize]{cleveref}  
\usepackage{enumitem}
  \newlist{inlinelist}{enumerate*}{1}
  \setlist*[inlinelist,1]{%
          label=(\roman*),
      }

\usepackage[normalem]{ulem}

\iftrue
 \newcommand{\comments}[1]{#1}
 \else
 \newcommand{\comments}[1]{}
 \fi

 \newcommand{\cut}[1]{\bgroup\renewcommand*{\cite}[1]{}\renewcommand*{\citet}[1]{}\comments{\textcolor{gray}{\sout{#1}}}\egroup}
\newcommand{\jlcut}[1]{\bgroup\renewcommand*{\cite}[1]{}\renewcommand*{\citet}[1]{}\comments{\textcolor{deepForestGreen}{\sout{#1}}}\egroup}
 \newcommand{\kscut}[1]{\bgroup\renewcommand*{\cite}[1]{}\renewcommand*{\citet}[1]{}\comments{\textcolor{orange}{\sout{#1}}}\egroup}
 \newcommand{\jgcut}[1]{\bgroup\renewcommand*{\cite}[1]{}\renewcommand*{\citet}[1]{}\comments{\textcolor{maroon}{\sout{#1}}}\egroup}

 \definecolor{maroon}{rgb}{.5,0,0}
 \definecolor{deepForestGreen}{rgb}{.15,.6,.04}
 \definecolor{magenta}{rgb}{0.7,0,1}
 \definecolor{orange}{rgb}{1,0.5,0}
 \definecolor{gray}{rgb}{.5,.5,.5}
 \definecolor{purple}{rgb}{0.4,0.1,0.6}

\begin{document}

\maketitle

\begin{abstract}
This paper pursues the insight that large language models (LLMs) trained to generate code can vastly improve the effectiveness of mutation operators applied to programs in genetic programming (GP).  Because such LLMs benefit from training data that includes sequential changes and modifications, they can approximate likely changes that humans would make.  To highlight the breadth of implications of such \emph{evolution through large models} (ELM), in the main experiment ELM combined with MAP-Elites generates hundreds of thousands of functional examples of Python programs that output working ambulating robots in the Sodarace domain, which the original LLM had never seen in pre-training.  These examples then help to bootstrap training a new conditional language model that can output the right walker for a particular terrain.  The ability to bootstrap new models that can output appropriate artifacts for a given context in a domain where zero training data was previously available carries implications for open-endedness, deep learning, and reinforcement learning.  These implications are explored here in depth in the hope of inspiring new directions of research now opened up by ELM.

\end{abstract}

\section{Introduction}

For many in the evolutionary computation (EC) community, the rise of deep learning (DL) has raised questions on its implications for EC.  Both approaches scale well with compute and both can yield useful discoveries and meaningful surprises.  Yet are they ultimately competing paradigms, or rather are they complementary?  In this paper we explore the latter possibility,
of considerable synergy, by highlighting an untapped implication of large language models (LLMs; \cite{bommasani2021opportunities,brown2020language}) for both genetic programming (GP; \cite{koza:book92,banzhaf1998genetic}) and open-endedness \cite{stanley:open,standish:ijcia03,bedau:alife00a}. 

In particular, in this new Evolution through Large Models (ELM) approach, an LLM trained on code can suggest mutations that are intelligent, thereby facilitating a dramatically more effective mutation operator that sidesteps many of the challenges that previously existed for evolving programs \cite{oneill:2010open}.  Interestingly, the benefits of ELM are also reciprocal back to deep learning: the set of samples generated through the LLM can eventually constitute a new training set in a novel domain that can then fine-tune the LLM to perform well in the new domain, a novel data-generation procedure. Furthermore, this approach ultimately opens up new opportunities in the pursuit of open-endedness by increasing the generative capabilities of the LLM solely through its own generated data.

LLMs have recently yielded impressive results in automated code generation \cite{chen:codex,li2022competition}. These models bootstrap from human knowledge by learning from very large datasets to achieve general coding competency.  The fact that such bootstrapping is possible is clearly relevant to GP.  After all, GP is in effect a generative approach to programming.  While it might seem at first glance that LLMs therefore might out-compete or subsume GP, in fact GP does still offer an advantage in situations where the particular class of programs targeted by the search is far (or even completely lacking) from the training distribution of the LLM.  In such cases, the LLM offers limited recourse (prompt engineering to learn an entirely new domain would be prohibitive), while GP can in principle evolve in any space (though in practice some spaces may be intractable due to the amount of mutation necessary to get consistent signal on fitness).

Interestingly (and perhaps surprisingly), the best of both worlds is easily attainable by combining them: simply by prompting the LLM to generate changes 
the LLM can serve as a highly sophisticated mutation operator embedded within an overarching evolutionary algorithm.  This way, the LLM in concert with evolution can steer each other towards the right region of the solution space even though neither evolution with a conventional mutation operator nor the LLM on its own could generate anything close.  
In effect, program evolution using LLM-based perturbation begins to bridge the divide between evolutionary algorithms and those that operate on the level of human ideas. That is, LLMs can be trained to approximate how humans intentionally change programs, while staying on the manifold of functionality. Furthermore, such LLMs can be further fine-tuned on successful perturbations for the purposes of self-improvement, culminating in a novel technique for iteratively enhancing the performance of ELM.

To highlight the potential of this approach, in this paper an entire dataset in a novel domain is generated from only a single mediocre starting example designed by hand by humans.  In particular, the domain is called Sodarace \cite{mcowan:sodarace,szerlip:iesor}, where two-dimensional
ambulating robots of arbitrary morphology are constructed for diverse
terrains.  Sodarace is cheap to simulate, allowing
fast iteration, and also makes it easy to appreciate the sophistication
of designs intuitively by simply watching the robot walk. In this way,
it facilitates quick assessment of whether a design is successful both
quantitatively and qualitatively.

To make the contribution of ELM explicit in the experiments in this paper, the Sodaracers are encoded as raw Python programs that output an enumeration of the ambulating robots' components.  That way, it is possible to demonstrate that ELM is a form of GP that can operate on a modern programming language directly, with no special provisions needed beyond the generic (i.e.\ \emph{not} previously trained in Sodarace) existing code-generating LLM.

A final important insight unlocked by this approach is that the ability to generate diverse solutions in a domain or part of the search space where there was little to no training data is foundational to bootstrapping an open-ended process \cite{bedau:open,standish:ijcia03,stanley:oreilly17}.  After all, open-endedness is fundamentally about searching outside the distribution of previous experience, which is exactly what ELM helps the LLM to do.  Because this novel capability has potentially far-reaching implications, we have chosen in this work to focus on the implications of the generated data that can be produced by ELM. Of course, ELM is applicable in many other contexts that will undoubtedly be explored in the future.

More specifically, experiments that follow show that generated data is sufficiently rich that it can serve as training data for fine-tuning LLMs to generate code for viable new Sodaracers consistently, and furthermore that reinforcement learning (RL) can even fine-tune an augmented LLM to output Sodaracers \emph{conditionally}, depending on the terrain.  In the future, such conditional invention has the potential to unlock entirely new kinds of open-ended processes, just as humans have open-endedly built civilization over centuries by conditionally inventing its constituents.

In short, the main contributions of this paper are (1) the ELM method for efficiently evolving programs through LLMs, (2) a technique for improving ELM's ability to search over time by fine-tuning its LLM-based mutation operator,
(3) a demonstration of ELM in a domain not included in the training data of the LLM, and (4) validation that data generated through ELM can bootstrap enhanced LLMs that offer a novel path towards open-endedness.

\section{Background}

This section reviews previous work in genetic programming, large language models, and open-endedness.

\subsection{Genetic Programming}

The field of genetic programming (GP) applies evolutionary algorithms to evolve computer programs
to solve problems \cite{koza:book92,langdon2013foundations,banzhaf1998genetic}. 
The promise of GP is that computer code is a computationally
universal representation that underlies much modern technology, including artificial intelligence itself. Therefore it is conceivable for GP to automatically evolve programs that achieve human-level (or beyond) performance across diverse application domains \cite{koza2006genetic}. However, there are obstacles in practice to its successful and widespread application to challenging problems.

One obstacle is that scaling GP to evolve increasingly complicated programs can be challenging \cite{oneill:2010open}, and that effectively applying GP to a new domain can require significant domain expertise. A researcher often must explicitly specify what functions, variables, and control structures are available to evolution \cite{koza:book92,brameier2001comparison}, which limits what ultimately can be evolved. In contrast, a human programmer can open-endedly decide what libraries to import and how to write many interdependent subroutines or classes. Research aims to lift these constraints, often through enabling modular reuse of code: e.g.\ through automatically defined functions \cite{koza:book92}, data-mining populations to find common sub-components \cite{jonyer2006improving}, or attempts to use solutions to previous problems when solving new ones \cite{seront1995external}. However, no method yet enables GP to scalably operate on human-designed programming languages with a minimum of domain-specific tweaking. 

A second obstacle is that nearly all GP methods explore through random perturbations of code, unlike humans, who through active practice improve their proficiency in making deliberate, complex, and coupled modifications to programs \cite{gugerty1986debugging,li2015makes}. Unlike perturbing e.g.\ neural network weights, wherein continuous parameters subject to small enough perturbations can predictably generate small changes in functionality \cite{lehman2018safe,schulman:trpo}, perturbing code requires discrete changes that often dramatically shift functionality \cite{galvan2010towards}, thereby complicating search. While there exist approaches towards more directed generation of offspring (e.g.\ building probabilistic models of high-performing programs \cite{salustowicz1997probabilistic}, evolving reproduction operators \cite{spector2002genetic}, or applying less-impactful mutation operators \cite{galvan2010towards}), the problem remains at core unsolved.

In contrast to GP, humans learn to reason about code in its full complexity through experimentation and learning. This iterative effort leaves a permanent signature in repositories of code, such as GitHub. The next section describes progress in training large language models upon such repositories as a potential way to bypass the above obstacles.

\subsection{Large Language Models}

Large language models (LLMs; \cite{bommasani2021opportunities,devlin2018bert,brown2020language}), trained on internet-scale data, have progressed at an
impressive pace in recent years. The main idea (in auto-regressive models such as GPT-3 \cite{brown2020language}) is to train increasingly-large 
neural networks (built on the popular transformer architecture \cite{vaswani:transformer}, sometimes with billions of parameters) on the seemingly simple task of next-token prediction (i.e.\ given a sequence of tokens seen so far, predict the proceeding token). Scaling such LLMs (and formulating problems of interest as natural language processing tasks) has resulted in 
groundbreaking performance across a wide range of tasks \cite{brown2020language,chowdhery2022palm}, 
including program synthesis \cite{chen:codex,hendrycks2021measuring,li2022competition}.

In particular, by training LLMs on large-scale coding data, e.g.\ from GitHub, 
it is possible to produce models with impressive
function-synthesis capabilities \cite{chen:codex,li2022competition}, highlighting the possibility to bootstrap the ability to fluently code from large-scale data. A further development are \emph{diff} models that are trained on diffs from GitHub \cite{diff_train}. A diff is an incremental change to a file that is committed to a version control system such as GitHub, accompanied by a \emph{commit message} describing the intent of the change. In this way, diff models are trained how, given a piece of code and any potential commit message, to suggest an informed \emph{change}. Through the lens of evolutionary algorithms, such diff models can be viewed as intelligent perturbation operators, providing a way to walk over the manifold of code (in a controllable way) through mimicking human programmers. An interesting further possibility is that such models are amenable to further
training through gradient descent, implying a potentially-powerful mechanism for self-adaptation (e.g.\ through reinforcing successful diffs during evolution). Both diff models and their capacity for self-adaptation are explored in this work as a way to improve GP.  However, it is also important to note that general language models not trained directly on diffs can also act in effect like diff models when given the right kinds of prompts (see Section \ref{background_diff}).

\subsection{Open-endedness}

With origins in the open-ended evolution community \cite{taylor:alife12,taylor2016open,bedau:open,standish:ijcia03} within artificial life, the field of open-endedness seeks to create algorithmic systems that produce never-ending innovation \cite{stanley:open}. Given the primacy of search with ML, research within open-endedness naturally has focused on refining algorithms for open-ended search, such as those driven by novelty \cite{lehman:ecj11,mouret:ec12} or curiosity \cite{pathak2017curiosity,stanton:curiosity}. While such focus has indeed lead to algorithmic progress, there is a growing awareness of the criticality of the \emph{environment} in which open-ended algorithms are applied \cite{grbic:evocraft,earle2021video,wang2020enhanced,soros:alife14}.

That is, the environment limits what can arise within the system and for how long its products can remain interesting. As a result, some have argued for more complex environments for open-endedness, such as video games \cite{grbic:evocraft,earle2021video}, and others have argued that features of the environment should co-evolve with agents \cite{wang2020enhanced,dennis2020emergent}. Yet a theory for what specific forms of additional such complexity is needed for enduring open-endedness has been lacking. This paper contributes a possible theory, arguing that agents
outputting inventions into the environment in response to previous inventions may be a principled
route to such continuing open-endedness.

One challenge in evolving aspects of the environment (such as inventions), is how they are encoded. Most research applies encodings that are specifically fit to describe some fixed part of a larger environment, e.g.\ a fixed way of describing edges within a maze \cite{brant:mcc}, or the shape of a 2-D landscape \cite{wang2020enhanced}. While sometimes the \emph{encodings} of these parts are universal (e.g.\ the CPPN encoding of landscapes in \cite{wang2020enhanced} can describe any landscape, and the RNN encoding of \citet{dennis2020emergent} can describe any maze), it is unclear how to expand the representation to include more of the environment without relying upon ad-hoc principles. This paper argues that computer programs are a general and powerful encoding for continually expanding the richness of an existing environment.

\section{Approach: Evolution through Large Models}

Three distinct components facilitate ELM.  First is the novel mutation operator driven by an LLM.  Second is an evolutionary outer loop that calls this mutation operator. Finally, the third component is a method for updating the LLM to improve based on its preceding performance.  Each of these is detailed in this section. 

\subsection{Mutation through Diff}

The main idea behind ELM centers on rethinking the mutation operator for code by exploiting the capabilities of LLMs.  In conventional GP, the language of the code and the types of changes allowed through mutation are both chosen intentionally to yield a reasonable chance that perturbations can lead to useful functional changes \cite{koza:book92}.  In contrast, LLMs unlock an entirely different basis for mutation: it would be more ideal if the mutation operator \emph{understood} the code and how it can be changed in interesting ways, more like a human than a stochastic event.  

LLMs can indeed be trained to output code in an autoregressive manner by exposing them to extensive programming examples \cite{chen:codex,li2022competition}.  
A \emph{diff model} \cite{diff_train} can similarly be autoregressively trained on a collection of code diffs (e.g.\ from GitHub).
Each diff targets a single file, where the 
file and diff are short enough to fit into the context of the LLM. 
The model is trained to predict the diff (formatted, for example, in unified diff format \cite{unified_diff}) from the concatenation of the file and the commit message, where the loss includes only the tokens that make up the diff, thereby encouraging the model to predict the diff but not to memorize the file and commit message. In other words, the model learns to predict plausible changes to code from examples of changes made to code by human programmers.
It is important to note that the idea of diff models (or their initial training) \cite{diff_train} is not a contribution of this paper, but diff models are rather a tool applied here in a new context (to produce mutations).

To achieve meaningful mutations, ELM can choose among a set of \emph{commit messages}, which convey to the LLM the details of the operation it should perform in lieu of mutation.  These messages offer significant power and nuance for calibrating mutation operators that is likely highly novel to anyone familiar with implementing mutation in GP or evolutionary algorithms in general.  In the experiment in this paper, the three commit messages and their respective probabilities of being chosen are:
\begin{itemize}
    \item \texttt{Changed make\_walker function.} (40\% chance)
    \item \texttt{Changed parameters in make\_walker function.} (30\% chance)
    \item \texttt{Small change to make\_walker function.} (30\% chance)
\end{itemize}
Of course, any commit message is conceivable. The LLM's ability to interpret general natural language means that the scope for research exploration (and domain-specificity) here is vast.

\begin{figure}[t]
\centering
\includegraphics[width=0.8\textwidth]{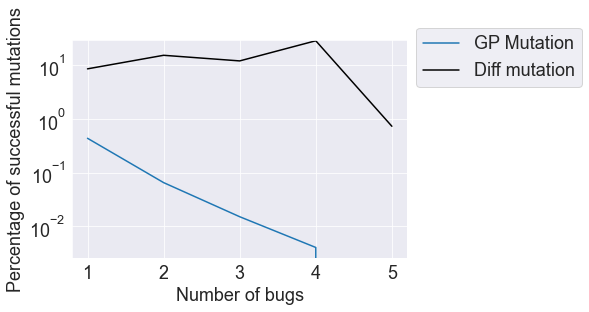}
    \caption{\textbf{Comparing diff mutation to GP mutation in fixing 4-Parity bugs.} 
    The figure shows how the ability of a single mutation to produce correct solutions changes as bugs are incrementally
    added to a working 4-Parity implementation. Note that success percentage is shown in \emph{log scale}, i.e.\ success for
    GP mutation decreases exponentially in the number of mutations (and produces no solutions when there are five bugs). 
    In contrast, diff mutation degrades only with the fifth bug. The conclusion is that LLM-based mutation can indeed make
    multiple sensible coupled changes to code.
    }
    \label{fig:intelligent_mutation}
\end{figure}

As a simple experiment to highlight diff models' ability to intelligently modify code, an implementation of a function  with an adjustable amount of bugs is perturbed with either a simple GP mutation operator or with
a 300M parameter diff model. The hypothesis is that an intelligent perturbation operator will be better able to make multiple correlated changes to code (in this case to correct the bugs). The \emph{4-Parity} task (which is inspired by a standard GP benchmark \cite{koza:book92}) serves as a representative test-bed. Note that a correct implementation of 4-Parity returns the sum of the four input bits, modulo two. Up to five bugs are introduced to 4-Parity, first by incrementally misnaming each of the variables in the sum calculation; and for the fifth bug, the modulo is changed from two to three. Then, perturbation operators are tested for their ability to (in one perturbation step) change the buggy version of the code to one that successfully passes unit tests. Results in figure \ref{fig:intelligent_mutation} highlight how with increasing bugs GP mutation becomes exponentially more unlikely to produce a successful solution (note that \emph{no} mutation from GP solves all five, given \num{100000} trials). In contrast, the diff operator is able to fix all five bugs, and its performance is impacted more by the number of \emph{different} types of bugs (i.e.\ the fifth bug affects the modulo calculation rather than renaming variables)  than by the raw number of bugs itself. Further details (including a supporting experiment with another task with similar results) are given in Appendix \ref{appendix:intelligent_perturbation}.

\begin{figure}[t]
\centering
\includegraphics[width=0.8\textwidth]{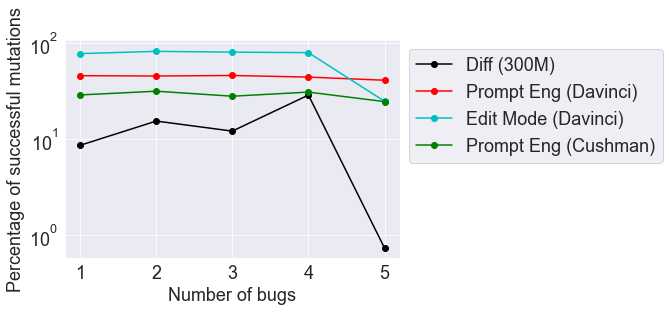}
    \caption{\textbf{Comparing alternate LLM-based mutations in fixing 4-Parity bugs.} 
    The performance of different mutation operators in fixing bugs is shown as bugs are incrementally added to a working 4-Parity implementation. Both edit mode and prompt-engineering approaches outperform the diff model applied in this paper's experiments. The conclusion is that both prompt-engineering on LLMs trained on code and using edit mode models from the OpenAI API are viable options to build upon the work in this paper.
    }
    \label{fig:api_parity}
\end{figure}

\label{background_diff}
Because the tools involved in an ELM implementation are unconventional, we finally wanted to highlight here several alternatives for implementing such systems in practice today.
One option is to use models available on the OpenAI API that can edit through following instructions \cite{edit_mode,ouyang2022training}. %
A second option is to create an intelligent mutation operator through few-shot prompting instead of through explicit training (as in the diff model). That is, one could design prompts for a model trained on code (like Codex \cite{chen:codex} or GPT-6-J \cite{gpt-j}). %
To show the potential to replicate (or improve upon) the results in this paper, we conducted a simple experiment comparing (on the 4-Parity problem) prompt engineering and edit mode to the diff model. Figure \ref{fig:api_parity} shows how models from the API outperform the diff model used in the paper. Further experimental details can be found in Appendix \ref{appendix:intelligent_perturbation}.

\subsection{The Evolutionary Algorithm and Implications for Open-Endedness}

Because the mutation operator is effectively a modular component for many evolutionary algorithms \cite{mitchell:book99,dejong:book06}, ELM can be implemented within a diversity of contexts.  Of course, the approach is most applicable to a case where the genetic encoding is through a known programming language, because that is how the benefits of the LLM will be realized.  Genetic encodings in natural language or any other language at which LLMs excel are also conceivable, but of course the utility of such encodings would depend on how they are applied and their mapping to a potentially useful phenotype. The experiments in this paper focus on Python 3 genotypes, which are also by their nature variable in length.  The ability to use modern programming languages as genotypes without the need for any special accommodation is a key benefit of ELM.

While there are many options for the evolutionary algorithm in the outer loop, we have chosen in this paper to implement ELM within a quality diversity (QD) algorithm \cite{pugh:frontiers16,mouret:arxiv15}.  An important motivation for this choice is that the emergence of the ability to search intelligently for arbitrarily complex programs is tantalizingly close to overcoming some of the key obstacles to open-endedness \cite{stanley:oreilly17}, and ELM is an opportunity to highlight this opportunity.

Recall that we do not yet know how to make an algorithm that
exhibits genuinely open-ended divergence. While there has been progress
towards open-endedness in recent years, the state of the art remains
\emph{weak open-endedness}, wherein novel and interesting discovery continues
only for a brief time, eventually ending in a plateau as the
possibilities are exhausted
\cite{brant:mcc,wang2020enhanced,wang:gecco19,brant:gecco20,stanley:open,baker2019emergent}.
In contrast, in \emph{strong open-endedness}, the process would never
plateau--if we left and returned a year later, or even a \emph{million} years
later, its products would continue to become more interesting over time. No algorithm
comes close to such an achievement, though it is evidently possible in
nature. 

The question then is what stands between today's algorithms and tractable strong
open-endedness.
This gap remains despite that recent work in
open-endedness appears to make progress. For example, the Enhanced POET
algorithm continues to generate diverse and increasingly complex
terrains for bipedal robots to solve \cite{wang2020enhanced}. In their
hide-and-seek experiment, \citet{baker2019emergent} show agents discovering
increasingly complex strategies like assembling blocks into a hideout.
Yet despite such algorithms clearly demonstrating the capability to
continue to invent new solutions, all such demonstrations share a
singular downfall: they slow down and eventually end.
Formalizing ELM within a QD framework in effect offers a novel opportunity to address this challenge.

This opportunity connects to the difficulty of formulating an artificial environment that imposes no limit on what even the most capable open-ended algorithm can achieve, as noted in the Background.
The challenge of devising artificial environments with unbounded potential
raises
the intriguing question of 
what property our universe and planet possess
that is lacking in current artificial environments. This question is
critically important for open-endedness because in the absence of that property, 
open-ended algorithms cannot demonstrate their full
potential.
If the problem indeed stems from the fact that artificial environments
to date offer only finite possible experiences until their potential is
exhausted, then to overcome this bottleneck the environment itself needs
to possess the potential to change forever.

Since the emergence of intelligence in nature, much environmental
change has been driven by the intelligent agents themselves.
Eventually, humans acquired the ability to leave
\emph{detached} artifacts in the environment that radically alter its
potential for themselves and other agents, like a house, a vehicle, or even a \emph{program}.
Unlike new organisms that are evolved over generations,
such \emph{detached conditional
things} (DCTs) are generated intentionally as a condition of the
observations of the agent. Once DCTs enter the world, open-endedness
accelerates because the environment is rapidly updating even within the
course of a single lifetime.

Each DCT creates an opportunity for further DCTs. For example, the
invention of the door creates the opportunity for keys to be invented,
which then sets the stage for lock picks, and so on. And because they
are detached, DCTs can leave a permanent legacy in the environment well
beyond the lifetime of their inventor. In this way, invention in the era
of DCTs is open-ended, and accordingly has continued for thousands of
years, from fire and wheels to space stations and computers.

This theory of DCTs supplies an abstract answer to the problem of a limited environment: Agents must be able to 
imprint the environment with DCTs in response to those already present within it.
However, realizing DCTs in practice requires addressing a separate question: how can agents be enabled to
efficiently invent DCTs of limitless complexity in a new domain? 

Interestingly, computer programs are universal representations, meaning that the procedure of assembling new artifacts can naturally be described algorithmically. For example, programmers have leveraged code to
help create enormously complex artifacts (like the layouts of computer chips or instructions for 3-D printers to produce complex physical objects).
Of course, programs themselves can function as DCTs.
In this way, a procedure that can search through modern program space
and ultimately generate such programs conditionally is a candidate for 
creating open-ended environments of unlimited capacity.  
The experiment in this paper will demonstrate in more detail how ELM makes 
such a construct conceivable; the significance of QD is that its ability to generate a diverse space of artifacts can serve as the bootstrap to obtaining agents capable of generating DCTs.  In short, the QD algorithm is generating \emph{training data} that can transform the LLM into a kind of DCT generator.

\begin{figure}[t]
\centering
    \includegraphics[width=\textwidth]{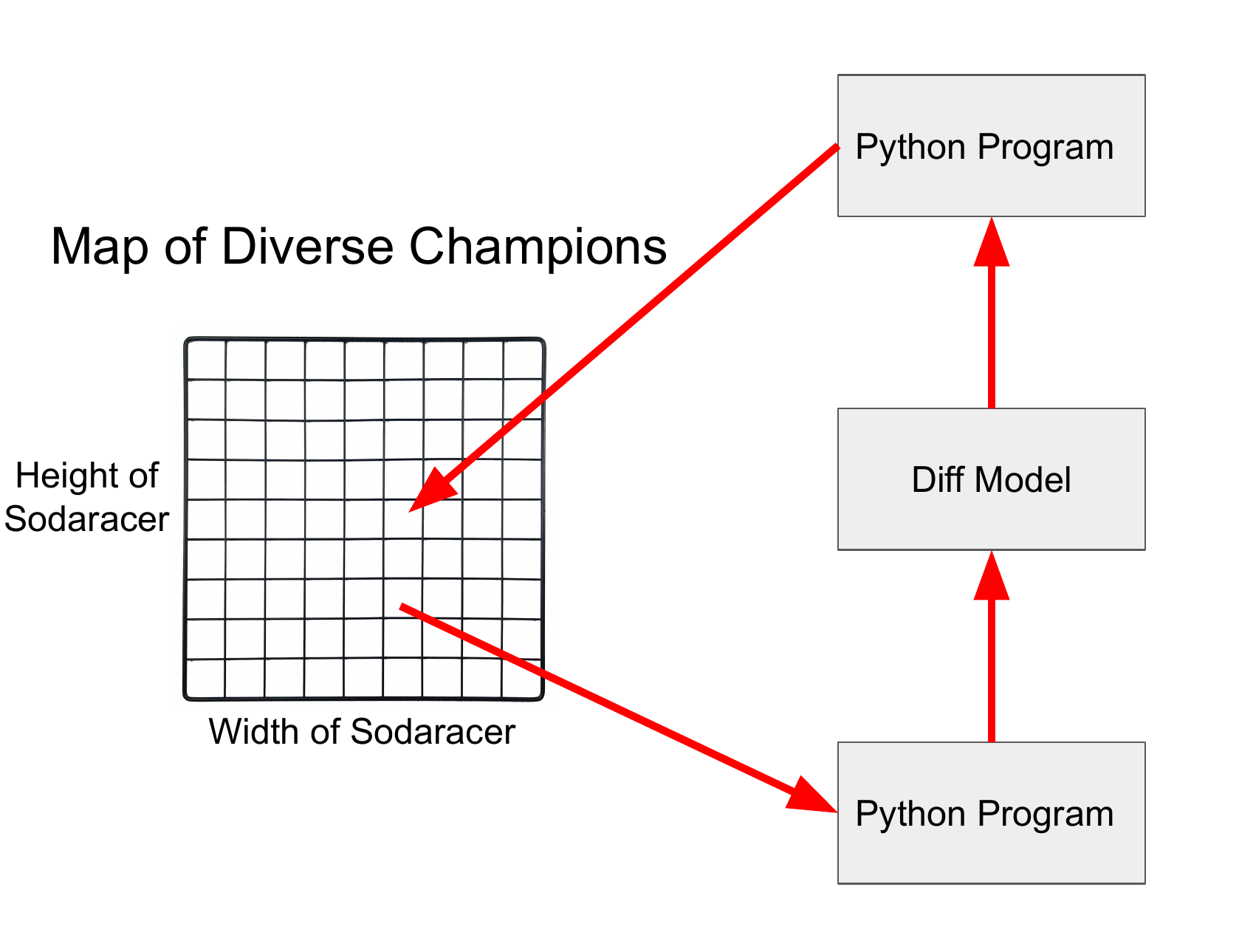}
    \caption{\textbf{MAP-Elites with ELM.} In each iteration, an existing Python solution is sampled from the map of solutions for each independent replica of a diff model. Each replica generates a batch of diffs that are applied to the sampled solution to generate modified candidate solutions. These candidates are evaluated and are then inserted into the map if they either establish a new niche or outperform the niche's current champion. Over iterations, a single initial seed program gives rise to a diversity of high-performing Python programs.}
    \label{fig:ELM-diagram}
\end{figure}

While any QD algorithm can work with ELM, the specific algorithm in the experiment in this paper is MAP-Elites \cite{mouret:arxiv15,cully:nature15} (\Cref{fig:ELM-diagram}).  
The core of MAP-Elites is a uniformly-spaced grid of niches (called the \emph{map}), that spans user-specified dimensions of solution diversity, called the \emph{behavior characterization}. 
Upon initialization, a single pre-existing (hand-designed in this paper) solution is evaluated and placed into the map. In each iteration thereafter, an inhabited niche is randomly chosen and the solution within that niche is perturbed by the diff model and evaluated. The new candidate solution is assigned its niche from its behavior characterization, and if that niche is unfilled or the new solution outperforms the niche's current inhabitant, it becomes the champion of that niche; otherwise, the candidate is discarded. In this way, over iterations of search, the map gradually fills with increasingly diverse and high-quality solutions.

\subsection{Fine-tuning the Diff Operator}

Interestingly, because the mutation (diff) operator is itself an LLM, it has the potential to be improved with respect to the domain.  While self-adaptation \cite{meyer2007self,kramer2010evolutionary,hansen2006cma} has a long tradition in evolutionary computation, including algorithms such as CMA-ES \cite{hansen2006cma} and natural evolution strategies \cite{wierstra2008natural}, the kinds of improvements possible in ELM are unique by offering the possibility of the LLM learning \emph{how to think about change}.
That is, ideas for changes that are most promising in one domain might be different than in another, and the richness of the LLM offers the potential to capture such nuance through experience.  In particular, the pre-trained diff model can be trained further (which is called \emph{fine-tuning}) with accepted diffs (by MAP-Elites) from initial iterations or runs of ELM.  That way, the diff operator updates to understand better the kinds of modifications that lead to either higher quality, more novelty, or both. 
This fine-tuning technique can cause ELM itself to improve over iterations.  Of course, over a long run, the ideal kinds of changes might change; continually fine-tuning based on recent experience can potentially track such drifting opportunities.  In this paper, the potential of fine-tuning is demonstrated through a single fine-tuning iteration, but the investigation of such continual refinement is an open research opportunity.
Note that the prompt-engineering approach to LLM mutation described at the end of Section \ref{background_diff} can also benefit from fine-tuning in this way.

\section{Experiment and Results}

The primary motivation for the experiment that follows is to give a taste of the breadth of implications of ELM, to evolutionary computation, to deep learning, and to open-endedness.  For this purpose, this experiment focuses 
on the problem of the \emph{invention} of complex artifacts (which could eventually serve as DCTs in a future more ambitious experiment).  
While the potential scope of applications for ELM is broad, the opportunity to learn to invent complex artifacts in an arbitrary domain extends directly from the augmented ability to search through programs; seeing this inventive capability realized thereby highlights novel opportunities opening up.  

The experiment will aim to
bootstrap from a few hand-written (and barely functional) examples of an invention 
into an LLM-based inventor that can fluidly output appropriate
inventions conditioned on its environment. This concept is demonstrated in the domain of \emph{Sodarace} \cite{szerlip:iesor,mcowan:sodarace},
a physics-based invention domain that serves as a cheap-to-simulate microcosm of invention. 
The goal in Sodarace is to construct from collections of masses and oscillating springs two-dimensional robots that can locomote competently. A wide range of interesting Sodaracer robots are possible, 
as highlighted by previous ML research \cite{szerlip:iesor} and the origins of the domain: Sodarace began as a web application called Sodaconstructor, wherein the human design of Sodaracers was sufficiently compelling for an online community to form around the endeavor \cite{mcowan:sodarace}. 

\begin{figure}[t]
\centering
\includegraphics[width=0.5\textwidth]{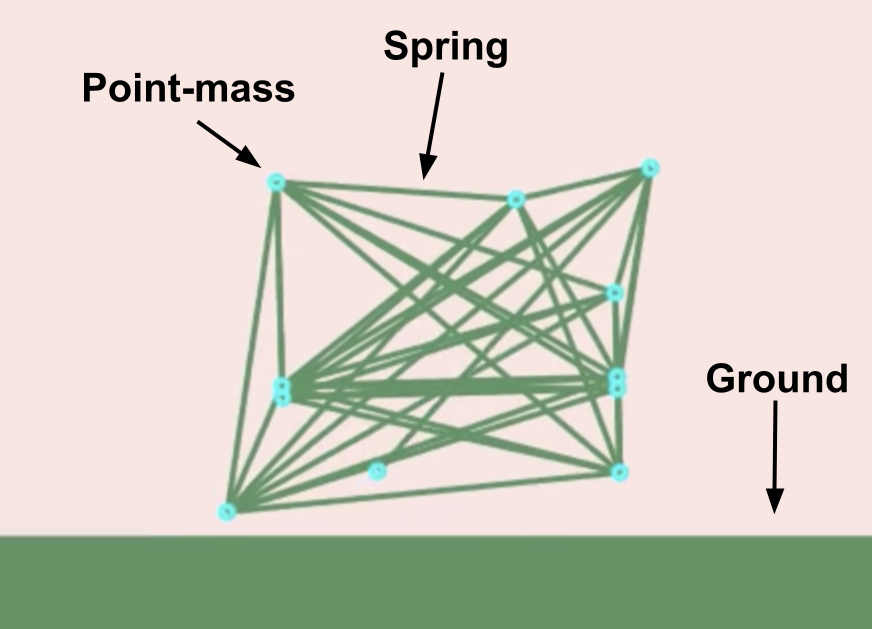}
    \caption{\textbf{An Example Sodaracer.} The objective in the Sodarace domain is to design a Sodaracer that locomotes effectively across the ground terrain. Labeled in the image are examples of a mass and a spring that connects two masses together. A Sodaracer design consists of a variable number of masses and springs, where springs also have oscillatory parameters that determine the Sodaracer's motion.}
    \label{fig:sodaracer}
\end{figure}

An individual Sodaracer (\Cref{fig:sodaracer}) is composed of a variable-sized collection of point masses (each fully-described by its initial 2-D position) and oscillating 
springs that connect masses together. The motion of the robot is driven by the oscillation of its springs, and each spring has parameters specifying the amplitude and phase of its oscillation (by convention all springs have the same period). To evaluate a particular Sodaracer, it is simulated in a specific terrain for a fixed amount of time 
and its ability to traverse that terrain is measured (i.e.\ how far the Sodaracer's center of mass moves along the x-axis); additionally, to measure the diversity of solutions for MAP-Elites,  features capturing gross aspects of the robot's morphology (i.e.\ its initial height, width, and total mass) are recorded. While a search algorithm could operate directly in the space of masses and springs, here instead 
LLMs output Python code that describes the morphology of the Sodaracer. For examples of such source code, see Appendix \ref{appendix:seed_source_code} and \ref{appendix:source_code}. In this way, the programs evolved by ELM are in effect \emph{indirect encodings} \cite{stanley:alife03,stanley:gpem07,bongard:gecco01,bentley:gecco99} for Sodaracers. That is, in principle any indirect encoding expressible through code could be evolved from scratch or modified by ELM.

More ambitiously than only generating a repertoire of Sodaracer designs, the experiment will attempt to implement an entire \emph{invention pipeline} that ultimately yields a novel kind of conditional LLM that can input a terrain and output an appropriate Sodaracer for that terrain.  ELM thereby serves as the initial \emph{data generation phase} of this pipeline, showing in this way how ELM can serve in general as a way of generating domain data for downstream deep learning where it did not previously exist.  Furthermore, in the future the ability to train such conditional inventors could serve as a foundation for an open-ended world of DCT-generating agents. 

In practice, the aim of the invention pipeline is to create an agent that can output complex artifacts conditionally, based on its observation of the environment.
If \emph{invention} is conceived as an action, then learning to invent conditionally can be viewed as a reinforcement learning (RL) problem
\cite{sutton:rlearn98}.  That is, for any given observation, the agent can be rewarded depending on the success of the resultant invention.  For example, in Sodarace, the agent might observe a specific terrain such as a hill and then output a design for a Sodaracer artifact.  The reward then depends upon the performance of the Sodaracer in the observed terrain.

This approach sounds straightforward--it is simply RL with complex outputs--but there is a problem.  If the agent has no prior experience in the domain (e.g.\ in Sodarace), then outputting even a valid (let alone working) artifact is effectively impossible.  As a result, there is no gradient for RL and it cannot bootstrap into the new domain.  

Therefore, to get RL started, some form of pretraining is necessary.  In effect, the RL fine-tuning described above is actually the last step in a \emph{pipeline}, where the preceding step is to teach the agent something preliminary about its domain.  For that purpose, an LLM can be trained on a large set of \emph{examples} from the target domain.  For example, these examples could be Sodarace walker designs.  After exposure to enough such designs, in principle the LLM knows something about the domain and can output sample artifacts from the training distribution.  With such knowledge later passed on to RL, it should now be possible to bootstrap into conditional fine-tuning.

However, there is \emph{still} a problem: where did all the examples come from for training the LLM?  If the hope is for the conditional inventor eventually to invent in a novel domain like Sodarace where a generic LLM  is unlikely to have any exposure, then the source for all the examples needed to train the LLM is itself elusive.  As a consequence, the pipeline needs yet one more preceding step--which is where ELM comes in--to generate a set of example artifacts from scratch, which could then train the LLM that will eventually bootstrap RL.

Generating a diverse and large set of initial training examples is a search problem.  However, because no LLM yet has any exposure to the right kind of data, it is a search problem within the invention space rather than within the weight space of neural networks.  Searching for diverse functional examples (to get a wide pre-training distribution) within the space of artifacts is the natural role of QD (i.e.\ MAP-Elites in this paper).  Combined with the diff function, the result is ELM, which yields a novel approach to generating training examples, thereby bootstrapping the entire process.

\begin{figure}[t]
\centering
\includegraphics[width=\textwidth]{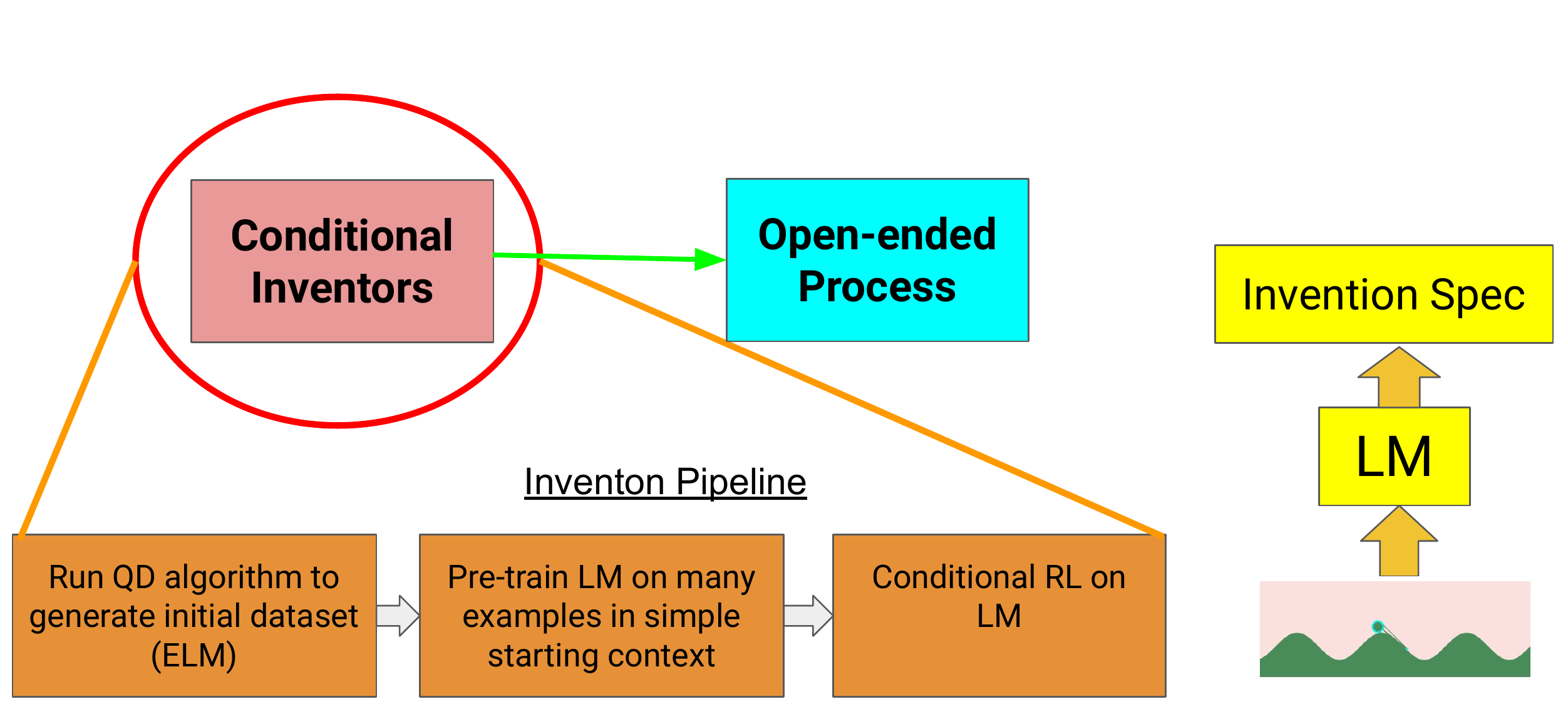}
    \caption{\textbf{The Invention Pipeline.} (left) A three-staged training pipeline bootstraps from a single example of an invention to an LLM that can output an invention tailored to its current condition. The hope for the future is for such a conditional inventor agent to help seed an open-ended process, wherein interactions between agents and their inventions spur continual innovation. (right) In the Sodarace domain, the conditional inventor observes the terrain, which conditions the LLM to output the specification of the desired invention.}
    \label{fig:method}
\end{figure}

To recap, what emerges is a three-stage \emph{invention pipeline} for training conditional inventors (\Cref{fig:method}):
\begin{enumerate}
\item \textbf{ELM.} Search for a diverse set of example artifacts (e.g.\ Sodaracers on flat ground).
\item \textbf{Pre-train the LLM with examples from ELM.} The LLM accordingly learns to output example inventions from the training distribution.
\item \textbf{Learn to invent conditionally.}
Splice new conditional inputs onto the LLM and fine tune it through RL to produce appropriate inventions for the conditions it observes.  
\end{enumerate}

\subsection{Encoding Sodaracers with Python}

Previous experiments targeting Sodarace have leveraged specialized evolutionary encodings \cite{szerlip:iesor}. Instead, in this work plain-text Python code acts as a generic representation for inventions. By showing how Python can be used to represent artifacts in an arbitrary domain, it opens up the possibility of using it as a generic encoding in diverse future domains.
More specifically, in the experiments an individual is evaluated by executing its code through the
Python interpreter. The product of the interpreter (for a viable individual) is a data structure containing the description of a Sodaracer (i.e.\ a Python dictionary containing lists of both point masses and springs), which can then be passed to the Sodarace simulator to evaluate the encoded Sodaracer's behavior. Note that Sodaracers are encoded in Python throughout the invention pipeline, i.e.\ ELM evolves Python programs and the language models in both latter stages of the pipeline are trained to output Python programs.

\begin{figure}[t]
    \centering
    \includegraphics[width=0.5\textwidth]{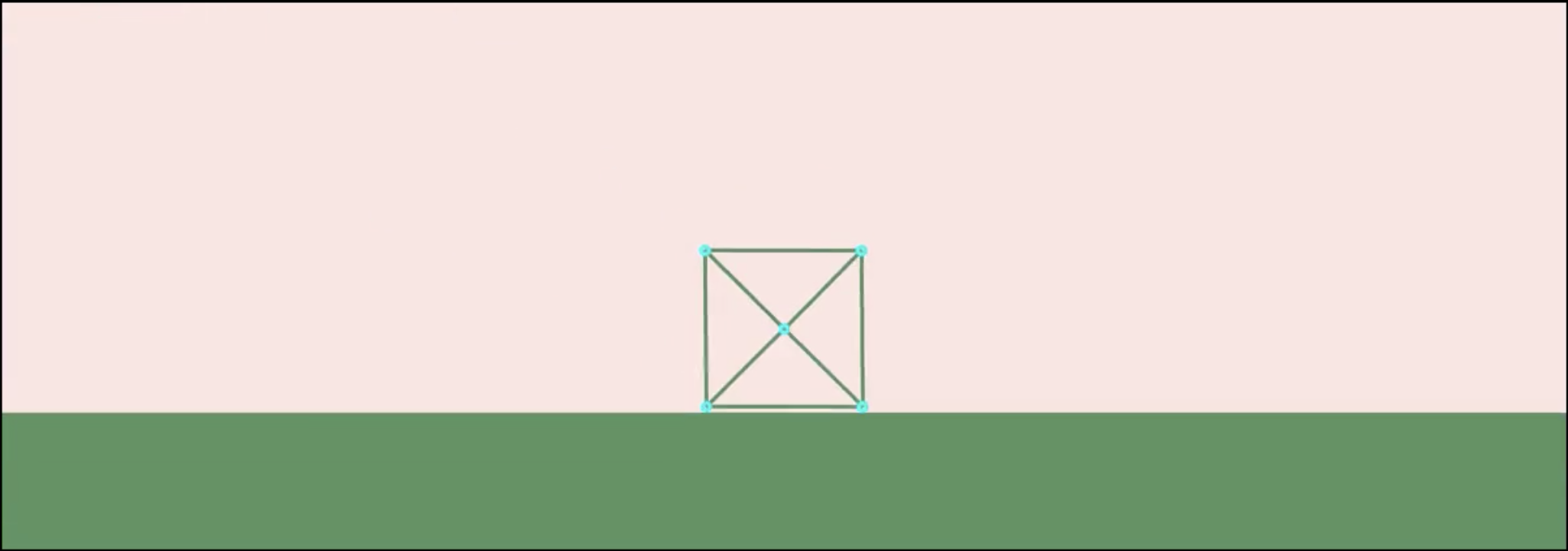}
    \caption{\textbf{Instantiation of a hand-designed square Sodaracer.} A video of this walker can be viewed at \url{https://y2u.be/jeP8Nsulu48}}
    \label{fig:videofig1}
\end{figure}

Preliminary experiments showed that the diff model's initial performance  (i.e.\ before fine-tuning) in creating useful perturbations depended upon the design of the ``interface'' through which Sodaracers were procedurally constructed. That is, while a Sodaracer can be constructed in Python by directly adding elements to a Python dictionary with keys such as ``joints'' and ``muscles,'' a more Pythonic interface (which was more effective and is what is used in the experiments) is to create a simple class with two methods: ``add\_joint'' (to add a spring) and ``add\_muscle'' (to add a point mass.) The idea is that such an interface (here encapsulated in a class called ``walker\_creator'') is closer to the training distribution of Python code (although still no Sodarace examples in this format exist). For example, below is the encoding of a simple hand-designed square Sodaracer (that is also used in the experiments as a seed), as well as its translation after being executed into a dictionary of joints and muscles. The interface also includes logic for ensuring that the Sodaracer will not break the underlying Box2D physics engine, e.g.\ that each joint is connected only to so many muscles, that the strength of muscles is limited, and that there is a minimum distance between joints.
Note that the benefit of evolving a program that produces a data structure rather than directly evolving the data structure itself relates to the benefits of indirect encoding (i.e.\ a program can leverage regularities through loops, conditionals, and functions, to efficiently encode large complex structures) \cite{stanley:alife03}. 
\Cref{fig:videofig1} shows an image of this walker when instantiated. 

\begin{listing}[!ht]
\begin{minted}[mathescape,
               linenos,
               numbersep=5pt,
               gobble=0,
               frame=lines,
               framesep=2mm]{python}
from walk_creator import walker_creator

def make_square(wc, x0, y0, x1, y1):
    """ Make a square with top left x0,y0 and top right x1,y1 """
    j0 = wc.add_joint(x0, y0)
    j1 = wc.add_joint(x0, y1)
    j2 = wc.add_joint(x1, y1)
    j3 = wc.add_joint(x1, y0)

    return j0, j1, j2, j3


def make_walker():
    wc = walker_creator()

    # the main body is a square
    sides = make_square(wc, 0, 0, 10, 10)
    center = wc.add_joint(5, 5)

    # connect the square with distance muscles
    for k in range(len(sides)-1):
        wc.add_muscle(sides[k], sides[k+1])
    wc.add_muscle(sides[3], sides[0])

    # one prong of the square is a distance muscle
    wc.add_muscle(sides[3], center)

    # the other prongs from the center of the square are active
    wc.add_muscle(sides[0], center, False, 5.0, 0.0)
    wc.add_muscle(sides[1], center, False, 10.0, 0.0)
    wc.add_muscle(sides[2], center, False, 2.0, 0.0)

    return wc.get_walker()
\end{minted}
\caption{Example Sodaracer-generating program.}
\label{listing:1}
\end{listing}

\begin{listing}[!ht]
\begin{minted}[mathescape,
               linenos,
               numbersep=5pt,
               gobble=0,
               frame=lines,
               framesep=2mm]{python}
{
    "joints": [(0, 0), (0, 10), (10, 10), (10, 0), (5, 5)],
    "muscles": [
        [0, 1, {"type": "distance"}],
        [1, 2, {"type": "distance"}],
        [2, 3, {"type": "distance"}],
        [3, 0, {"type": "distance"}],
        [3, 4, {"type": "distance"}],
        [0, 4, {"type": "muscle", "amplitude": 2.12, "phase": 0.0}],
        [1, 4, {"type": "muscle", "amplitude": 2.12, "phase": 0.0}],
        [2, 4, {"type": "muscle", "amplitude": 2.12, "phase": 0.0}],
    ],
}
\end{minted}
\caption{Intermediate Sodaracer representation from running the above Python seed program.}
\label{listing:2}
\end{listing}

\section{Pipeline Stage 1: Data Generation through ELM}

Recall that the aim in this first stage is to generate a large variety of high-quality 
examples 
from a single example starter program through ELM. In this stage of the pipeline, the  Sodarace environment is a simple flat terrain. 

Recall that ELM in this experiment will evolve through MAP-Elites (\Cref{fig:ELM-diagram}) \cite{mouret:arxiv15}. The core of MAP-Elites is a uniformly-spaced grid of niches (called the \emph{map}), that spans user-specified dimensions of solution diversity, called the \emph{behavior characterization}. In experiments here, the behavior characterization consists of the height, width, and mass of Sodaracers, and the map is a $12\times12\times12$ grid into which any behavioral characterization can be mapped. 
Upon initialization, a single hand-designed solution is evaluated and placed into the map. In each iteration thereafter, an inhabited niche is randomly chosen and the solution within that niche is perturbed by the diff model and evaluated. The new candidate solution is assigned its niche from its behavior characterization, and if that niche is unfilled or the new solution outperforms the niche's current inhabitant, it becomes  the champion of that niche; otherwise, the candidate is discarded. In this way, over iterations of search, the map gradually fills with increasingly diverse and high-quality solutions.

\begin{figure}[t]
    \centering
    \includegraphics[width=0.4\textwidth]{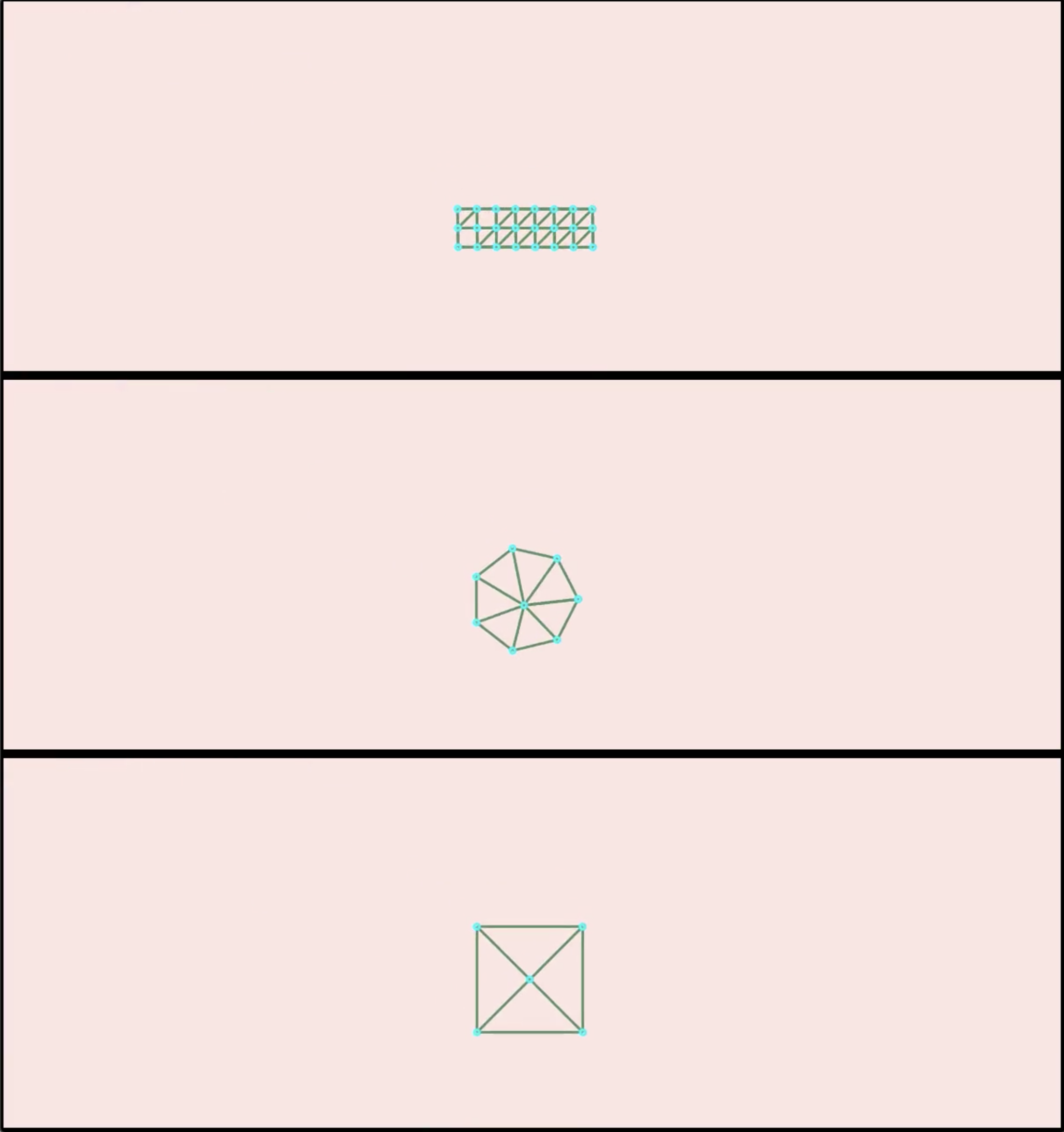}
    \caption{\textbf{The three seed solutions.} From top to bottom: CPPN seed, radial seed, and square seed. A video of these walkers is at \url{https://y2u.be/jeP8Nsulu48} (same video as for Figure \ref{fig:videofig1}).}
    \label{fig:videofig1-b}
\end{figure}

To address the need for seed solutions, four simple seeds were written that explore different architectural motifs: the Square seed, the Radial seed, and two CPPN-like seeds (CPPN stands for \emph{compositional pattern-producing network} \cite{stanley:gpem07}); note that source code for these seeds is provided in Appendix \ref{appendix:seed_source_code}.
The Square seed instantiates a polygon-like bias, by including a function that creates a square composed of four masses from two coordinates, and code that calls that function and connects the masses together with a for-loop. The Radial seed instead implements a radial bias by replacing the square-generating function with a function that places a given number of masses in a circular shape. Finally, the CPPN-like seeds roughly implement the CPPN-based encoding used by previous work in Sodarace \cite{szerlip:iesor}, i.e.\ it places masses and connects springs between them as a mathematical function of their coordinates. 
The CPPN-based seed's code can be neatly divided into (1) implementing the core functionality of a CPPN, and (2) describing a particular instantiation of a CPPN, and thus enables exploring the consequences of letting core functionality of the encoding itself evolve. To this end, there are two CPPN seeds, one in which the CPPN encoding is fixed, called the CPPN-Fixed seed, and one where it is mutable, called the CPPN-Mutable seed.
Note that these seed programs were not highly-tuned as the videos in \Cref{fig:videofig1-b} highlight.

\subsection{Experimental Details and Results}

Three independent runs of ELM were conducted with each seed, running for \num{1024000} evaluations each (composed of 2{,}000 iterations of $512$ diffs per iteration). A 300M parameter pretrained diff model \cite{diff_train} served as the perturbation operator in these experiments.

\begin{figure}[t]
\centering
\includegraphics[width=0.85\textwidth]{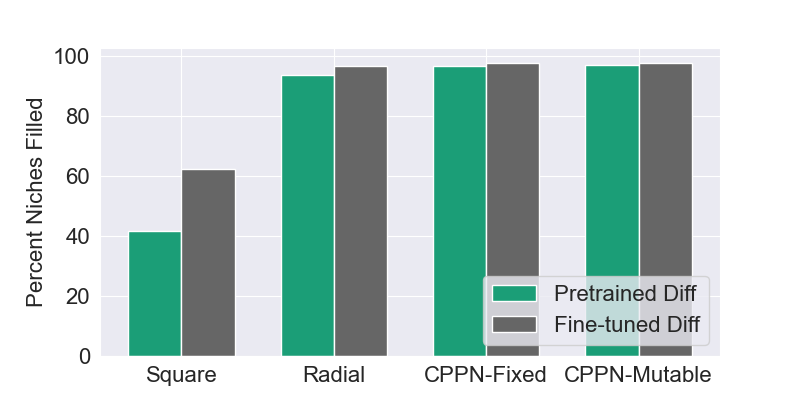}
    \caption{\textbf{Amount of niches filled across seeds.} 
    The figure shows the percentage of all niches (1{,}728 in total) that are filled by the end of ELM runs across different seeds. Results are averaged across three independent runs for each seed. In general, nearly all seeds fill the map, although the Square seed proceeds more slowly than other seeds.}
    \label{fig:stage1_niches}
\end{figure}

One metric of success for ELM is the number of niches filled, which represents the diversity of data generated by ELM, under the hypothesis that such diverse data will benefit later pipeline stages. \Cref{fig:stage1_niches} shows that runs of ELM tend to discover a large proportion of niches,
highlighting how the system can bootstrap from a single user-provided example to fill the space of desired possibilities. However, the speed of spreading through niches varies across seeds; in particular, introducing loops and/or function composition is required for the Square seed to spread into high-mass niches (e.g.\ to connect many squares together), which emerges slowly in some runs.

\begin{figure}[t]
\centering
\includegraphics[width=0.85\textwidth]{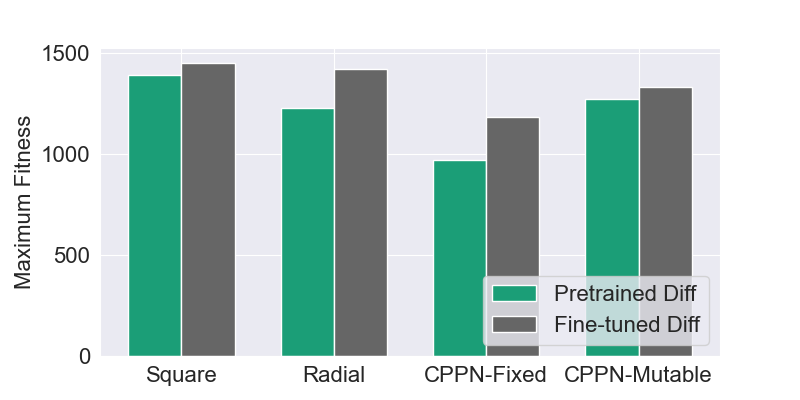}
\caption{\textbf{Maximum fitness across seeds.} The maximum performance attained on average by different seeds is shown. The results suggest that ELM's capacity to find high-fitness solutions is somewhat robust to seed design.}
\label{fig:stage1_maxfit}
\end{figure}

\begin{figure}[t]
\centering
\includegraphics[width=0.85\textwidth]{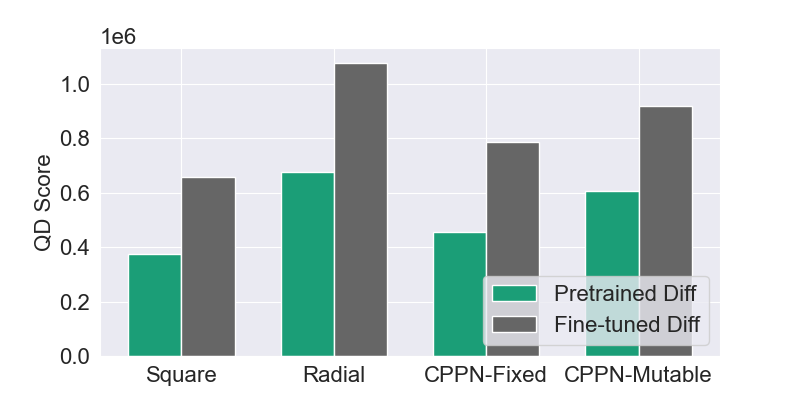}
\caption{\textbf{Quality diversity score across seeds.} Shown is the average final QD score attained by runs initialized from different seeds. The conclusion is that fine-tuning the diff model has a significant impact on attained QD score, as does the choice of seed.}
\label{fig:stage1_qd}
\end{figure}

Beyond diversity, the quality of solutions is also important. A gross measure of quality is the maximum fitness discovered by runs, shown in \Cref{fig:stage1_maxfit}. A more nuanced metric that takes both quality and diversity into account is the QD score \cite{pugh:frontiers16}, calculated as the sum of the performance of all champions in the final map. This metric, shown averaged over runs in \Cref{fig:stage1_qd}, rewards both quality (having higher scores in each niche) and diversity (having discovered more niches), and thus serves as a succinct measure of ELM's goal of accumulating diverse, high-quality solutions (and in later stages in the pipeline, of how well an LLM has modeled the distribution
of solutions that ELM has uncovered). 
Attainment of QD differs across seeds; while the CPPN seed uncovers diversity most quickly, the Radial seed generates higher-quality solutions on average. The relationship between the seed and the products of search is complex and deserves further future study (see also Appendix \ref{appendix:robustness} for further analysis of seed robustness). 

\begin{figure}
\centering

    \includegraphics[width=.7\textwidth]{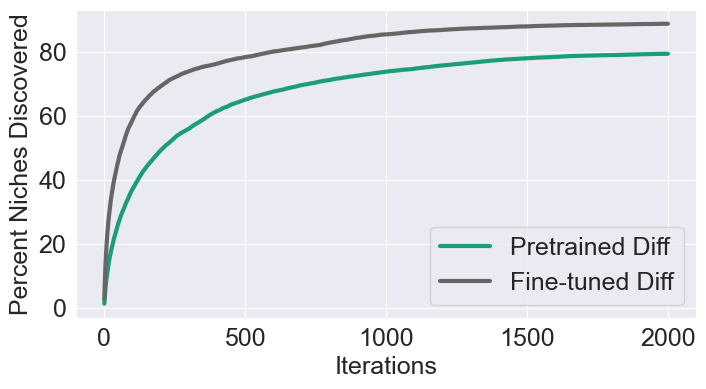}  \\
    \textbf{(a) Niches Reached}  \\

    \vspace{0.1in}
    \includegraphics[width=.7\textwidth]{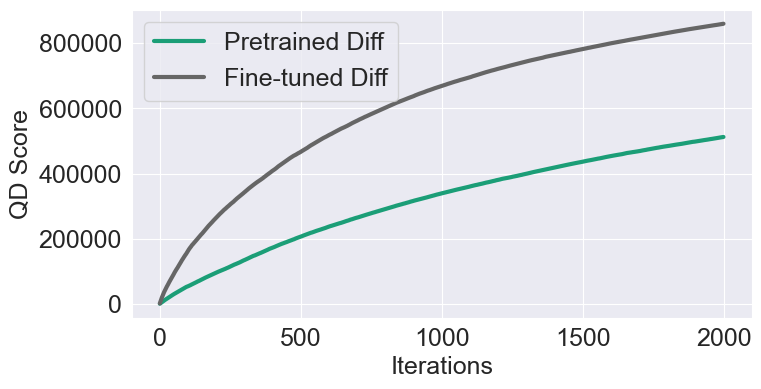} \\
    \textbf{(b) QD Score} \\

    \vspace{0.1in}
    \includegraphics[width=.7\textwidth]{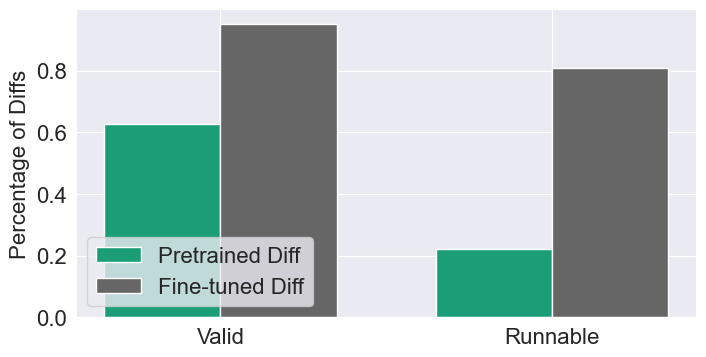} \\
    \textbf{(c) Diff Quality} \\

    \caption{\textbf{The impact of fine-tuning the diff model on the performance of ELM.} For both the pretrained diff model and the fine-tuned one, shown are (a) the number of niches reached, (b) QD score of the produced map, and (c) percentage of valid/runnable diffs proposed.
    The experiments demonstrate that fine-tuning the diff model improves performance of the evolutionary process across all three metrics.}
    \label{fig:finetuned-vs-pretrained-dff}
\end{figure}

Fine-tuning the diff model on accepted diffs from
an initial series of runs greatly increased performance (\Cref{fig:finetuned-vs-pretrained-dff}); while Sodarace-generating programs are out-of-distribution for the pretrained diff model (applying a Python encoding to this domain is a novel enterprise), fine-tuning effectively aligns the diff model with the domain, an interesting result. \Cref{fig:finetuned-vs-pretrained-dff}c
shows how 
 the fine-tuned diff 
 model produces a significantly higher percentage of diffs that are valid (i.e.\ able to be applied) and runnable (i.e.\ the patched program is executable). Because of their higher performance, the output of runs applying the fine-tuned diff model are the ones passed to later stages in the pipeline.
 
Note that further rounds of fine-tuning are possible (e.g.\ fine-tuning the diff model again from the improved products of the first round); however preliminary experiments showed diminishing returns. Future work could explore how to continually improve such models, such as by identifying and encouraging more impactful perturbations instead of including and weighting equally all accepted diffs. 

The seeds and fine-tuned diff model also qualitatively impact the kinds of solutions discovered by ELM. While the Radial seed performs well quantitatively (in terms of quality and diversity), it turns out that its products tend to exploit chaotic dynamics that seem overfit to the flat terrain (this hypothesis is tentatively validated in the Stage 3 experiments). The Square and CPPN seeds in contrast are more likely to output inventions that leverage more predictable dynamics. 
For these reasons, the Radial seed runs were not ultimately used in future stages.

A video selection of the highest-quality Sodaracers from these initial runs
that showcases the considerable diversity uncovered
can be viewed at \url{https://y2u.be/QNyNtvwA9FI}. 
An example of a lineage of Sodaracers progressing from the Square seed to a high-quality final Sodaracer can be seen at \url{https://y2u.be/M9pAJuX6dyM}.
In short, ELM shows that by combining the an intelligent LLM-based mutation operator with a QD algorithm it is possible to generate hundreds of thousands of working training examples in a completely novel domain where no data was previously available.

\section{Pipeline Stage 2: Language Model Training}
\label{sec:pipeline-stage-2}

The product of Stage 1 is a collection of programs, whereas Stage 3 RL requires an initial model that can output valid Sodaracer-generating programs. Thus, the second stage of the invention pipeline fine-tunes an LLM on the products of ELM, which serves as the initialization for an RL-based conditional inventor. To do so first requires compiling the results of Stage 1 into a fine-tuning dataset.

While there are many ways to distill a dataset of programs from runs of ELM, a simple thresholded approach is adopted here (although see Appendix \ref{appendix:finalmap} for another simple approach that did not work in practice). The main idea is to append all reasonably-capable solutions for each niche.

In more detail, from each run all solutions ever admitted to the map are included, subject to meeting a minimal bar for performance. Some parts of the behavior space offer more stringent challenges (i.e.\ it is more difficult to locomote when required to be tall but not wide and to have low mass), and yet in some terrains encountered in Stage 3, those kinds of solutions may yet be most effective despite their low level of absolute performance. Thus, for each niche, the maximum performance across all runs is calculated, and the minimal bar for inclusion is set as a percentage of that per-niche score. With a higher percentage threshold, less data is included, but the quality of that data will be higher.

As noted in the previous section, solutions from the Radial seed were qualitatively chaotic. Furthermore, preliminary experiments suggest that such chaotic behavior significantly harms downstream Stage 3 performance. For these reasons Radial runs of ELM were excluded from the LLM datasets. Datasets for each of the remaining treatments were compiled from 9 runs from ELM with the fine-tuned diff model (3 runs for each of the Square, CPPN-Fixed, and CPPN-Mutable seeds). In total, the 50\% cut-off threshold dataset consisted of 280K examples, and the 80\% cut-off threshold dataset contained a subset of 95K of those examples.

A variety of pretrained code-generating models were then fine-tuned with these examples (using the standard LLM log-probability loss), leaving out 5\% of the data to serve as a test set. Models ranging from 0.1M to 680M parameters were trained (architectural details for these models can be seen in Appendix \ref{appendix:arch}).
Also, as a control to support the hypothesis that Sodarace models benefit from code-generation pretraining, a 300M model was also trained instead from a random initialization (signified with ``RI'' in charts that follow).

\begin{figure}[t]
\centering
\includegraphics[width=0.75\textwidth]{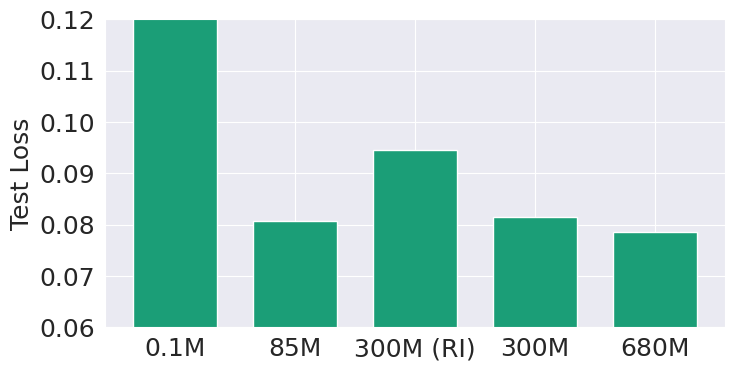}
\caption{\textbf{Test loss across model sizes.} The minimum test loss achieved by training runs on the 80\% Percentage Threshold dataset across model sizes is shown. Model sizes above 85M may not better-fit the data, and random initialization hurts performance.}
\label{fig:stage2_testloss}
\end{figure}

Minimum test-losses (i.e.\ loss on generated Sodaracers held-out from the fine-tuning dataset) of the 80\% Percentage Threshold models are shown in Figure \ref{fig:stage2_testloss}. The 50\% Percentage Threshold models exhibit qualitatively similar results across model size (but as both thresholds represent different datasets, loss values are not directly comparable between them). The conclusions are that model sizes above 85M may not better fit the data, and that random initialization does hurt performance relative to fine-tuning from a model pretrained on code.

\begin{figure}
\centering
\includegraphics[width=1.0\textwidth]{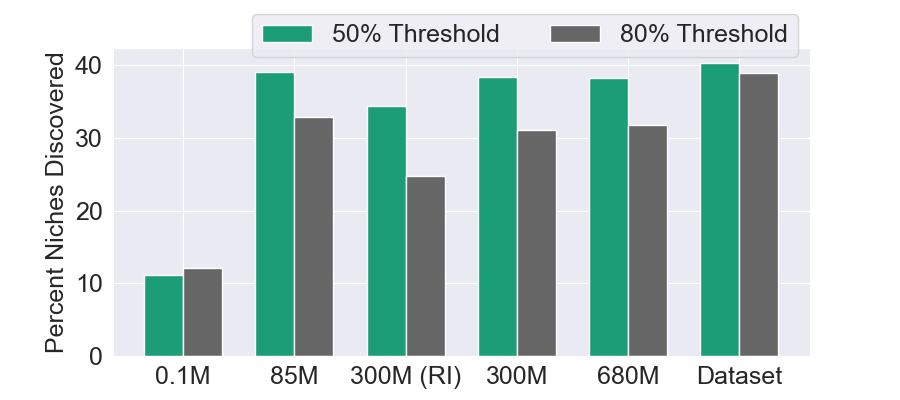} \\
(a) Number of Niches Filled

\includegraphics[width=1.0\textwidth]{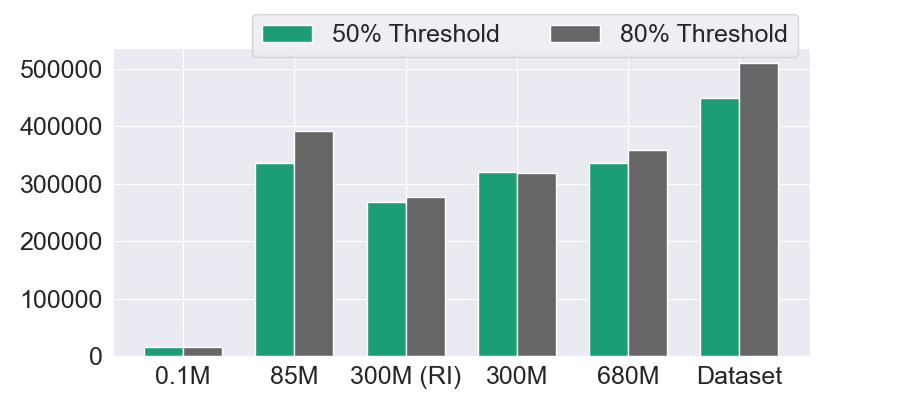} \\ (b) QD Score

\caption{\textbf{Measuring the quality and diversity of model samples.} Two metrics evaluating samples from trained LLMs are shown (across model size and training dataset): (a) the percentage of niches discovered  and (b) the QD score achieved. The 80\% threshold dataset is on average less diverse but of higher quality than the 50\% threshold dataset, and induces the same properties in models trained upon it. There is not a trend in increasing quality or diversity as model size increases beyond 85M, and random initialization hurts performance.
	}
\label{fig:stage2_samples}
\end{figure}

However, loss is not the whole story. The interesting question for Stage 2 is whether the LLMs trained from the data generated in Stage 1 can generate the same diversity and quality of data. Therefore, the QD score metric and number of niches discovered (both of which were also reported for Stage 1) are calculated for samples taken from trained LLMs. Because these metrics can be maximized by a model that memorizes the data, and because empirically QD score was more correlated with loss on the training set rather than the test set, the LLM checkpoint for each model is selected on the basis of lowest training loss. In particular, 1{,}024 samples are taken from each model, which are then evaluated and inserted into a new MAP-Elites map. For comparison, the same metrics are calculated using the Stage 1 dataset, by taking the same number of samples from it and evaluating them in the same way. These results are shown in \Cref{fig:stage2_samples}, highlighting that the model samples achieve a similar level of performance as dataset samples, suggesting that they have modeled the data well. Also, there is a slight but consistent QD benefit from models trained on the 80\% cutoff dataset, reflecting the higher average QD of that dataset.

\begin{figure}[t]
    \centering
    \includegraphics[width=0.5\textwidth]{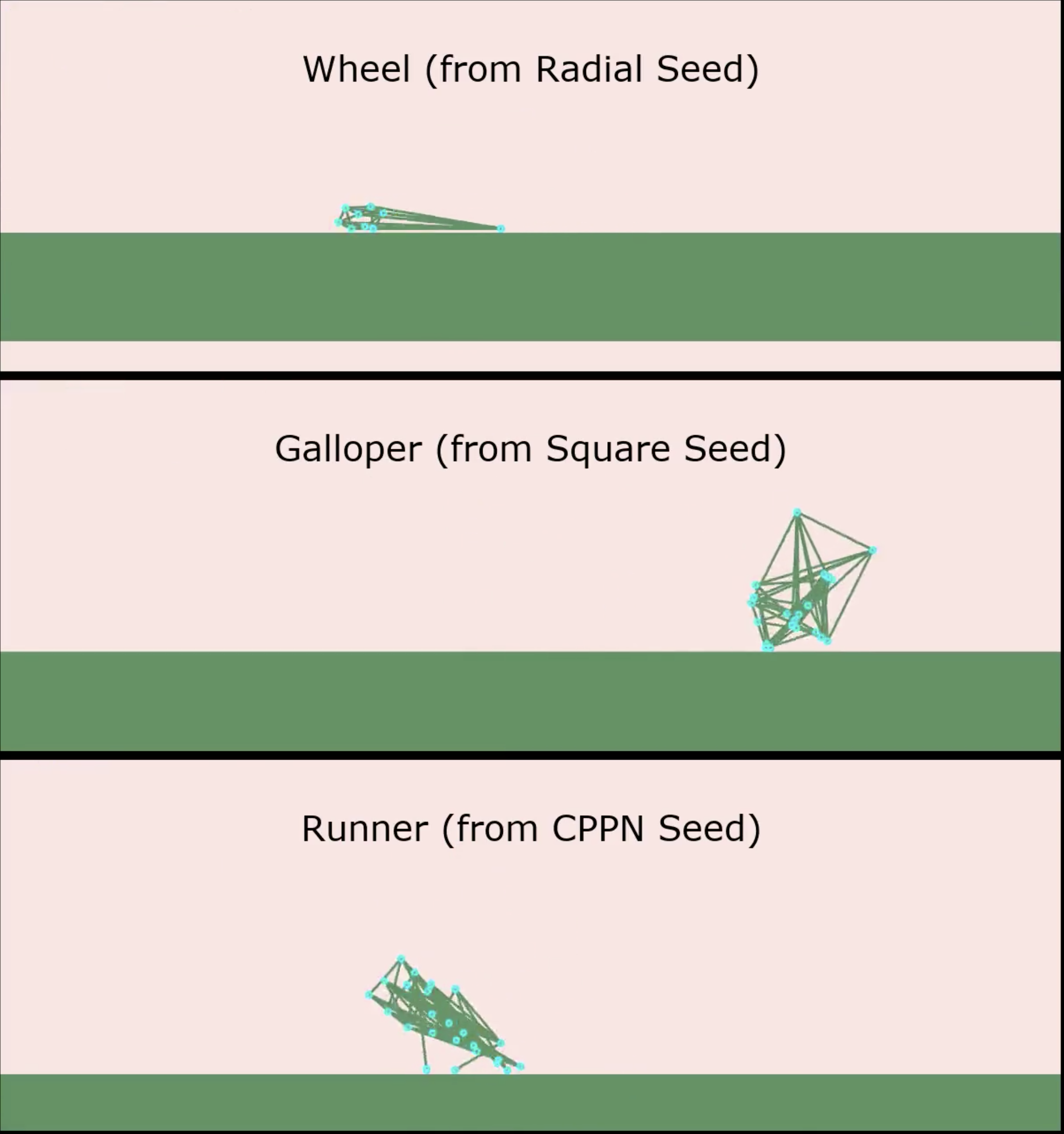}
    \caption{\textbf{Generalization tests.}
    In this test, the model is asked to complete samples, taken from the first half of the dataset from the unseen runs.
    These unseen originals are shown in the videos at \url{https://y2u.be/8C2K5fk28HI}. From top to bottom: Wheel, from radial seed; Galloper, from square seed; Runner, from CPPN seed.}
    \label{fig:videofig4}
\end{figure}

A natural further question is how well the model will do when taken out of distribution, i.e.\ how well has it really internalized the dynamics of Sodarace? That is, the training and test set for fine-tuning are taken from the same runs, and thus the model will likely have encountered all of the motifs in the test set, and so it may not be a representative test of how well the model will generalize in the future. A preliminary test in this spirit is to take the first half of the Python programs describing several inventions from unseen runs, and explore the capacity of different models to generate functional completions. Though the Radial seed usually produced chaotic Sodaracers, in one preliminary run of ELM with the Radial seed, a functional wheel was discovered. As noted previously data from this run (or any other radial runs) was not used to train the models in Stage 2, nor was it used to fine-tune the diff model in Stage 1; thus the ability to complete the wheel can serve as a proxy for generalization. Similarly, two other high-performing individuals were taken from other preliminary runs of the CPPN seed and the Square seed, to create a set of three out-of-distribution completion tests. See \Cref{fig:videofig4} for visualizations of these walkers, including videos; source code for these generalization examples can be found in Appendix \ref{appendix:completion_targets}). Note that further tests of generalization are documented in Appendix \ref{appendix:probing}.

\begin{figure}[t]
\centering
\includegraphics[width=0.75\textwidth]{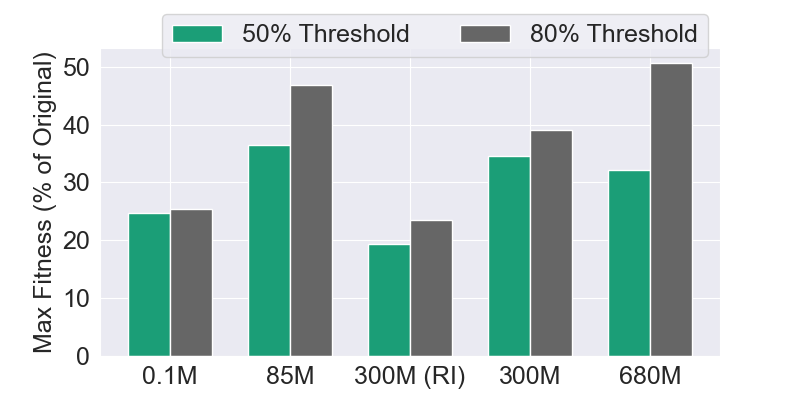}
\caption{\textbf{Out of distribution completion performance.} Shown is the percentage of the original solutions' performance that is attained by completions from trained LLMs. The percentage shown is the maximum attained over 1{,}024 independent completion samples from each model. The results are averaged over three out-of-distribution solutions (taken from runs not included in LLM training). The conclusion is that the 80\% threshold models perform better than the 50\% threshold, and that there is no obvious trend in performance once model size reaches 85M parameters.
}
\label{fig:stage2_ood}
\end{figure}

For each of the three completion tasks, 1{,}024 completion samples are taken from each model and then evaluated in simulation. In contrast to the in-distribution metrics, in this generalization-focused test, performance was more correlated with the model’s test loss rather than training loss, and thus what checkpoint to evaluate for each model was selected on the basis of lowest test loss. Results are shown in \Cref{fig:stage2_ood}, highlighting that larger models, and those trained on the 80\% threshold, generally perform better at this task. Note that the randomly-initialized (RI) 300M model significantly underperforms, providing more evidence that pretraining on code provides a valuable prior.

Videos of the best-performing sample for the Wheel completion from each model are at \url{https://y2u.be/-LW2cCwSdRU} (for the 80\% threshold dataset; the random-initialized 300M model is not shown because it generated no valid samples for this completion). For the Galloper and Runner completions, the structure and/or behavior of completions often does not match the original sample (especially for the Galloper). In the following linked video, a higher-performing completion is shown for both of the Galloper and the Runner: \url{https://y2u.be/XR3L4cZ83xU}.

Overall, these results show that an LLM can effectively integrate synthetic data generated through ELM in a novel domain.

\section{Pipeline Stage 3: Conditional RL}
\label{sec:pipeline-stage-3}
In the final stage, reinforcement learning (RL) is invoked to fine-tune the pretrained LLM output by Stage 2 of the pipeline. The goal is to produce a model that outputs Python programs representing Sodaracers in response to \emph{particular terrains}. Importantly, the output of Stage 2 is an \emph{unconditional} model, in the sense that it samples Sodaracers from a distribution defined by the output of Stage 1, without considering the terrain in which the samples will be deployed.  The first step in Stage 3 is thus to convert the model to a \emph{conditional} one, i.e.~a model that accepts terrains as inputs, and produces samples of Sodaracers in response.

To achieve this functional form, we first introduce the notion of a \emph{terrain embedding network} (TEN). The role of the TEN is to map a representation of the terrain to a representation that can be used by the model to sample conditionally. In particular, the output of TENs is a vector (or sequence of vectors) in $d$, the dimension in which the model embeds tokens. That way, the output of the TEN can be treated as the activation from a given prefix, and the model can proceed in effect now sampling conditioned on the output of the TEN.

Concretely, an unconditional autoregressive LLM defines a sampling distribution over a sequence of tokens $\vx = (x_1, \hdots, x_n)$ as $p_\vtheta(\vx) = \prod_{i=1}^{n} p_\vtheta(x_i | x_{<i})$. In this stage, we introduce the additional module $f_{\text{TEN}}$, which represents terrains $t$ in $\sR^d$. As $f_{\text{TEN}}(t) \in \sR^d$, we can consider the resulting conditional model without further modification: 
\begin{equation}
\label{eqn:conditionalLM}
    p_\vtheta(\vx | t) = \prod_{i=1}^{n} p_\vtheta \left( x_i | x_{<i}, f_{\text{TEN}}(t) \right). 
\end{equation}
This approach is similar to the controllable transformer proposed by \citet{keskar2019ctrl}, but with the conditional codes being the output of a TEN, rather than particular tokens from the existing vocabulary. 

Given a distribution over terrains $p(t)$, an RL setting is constructed to train the parameters of the TEN and further finetune the LLM parameters to the conditional setting. In particular, an episode now consists of sampling $t \sim p(t)$, and sampling a program from the conditional distribution defined in \Cref{eqn:conditionalLM}. The program is converted to a Sodaracer, evaluated in simulation with the terrain $t$, and the reward is defined as the absolute distance traversed by the Sodaracer in a given period of time.

\subsection{Terrain Distributions}
\label{sec:terrain-distributions}
In this experiment, the distribution over terrains that the model is exposed to is chosen to explore the viability of producing conditional inventors with the Invention Pipeline. The future vision is to lay the groundwork for the ability to deploy agents capable of conditional invention in rich, potentially multi-agent environments that support the development of open-ended processes. In such settings, it stands to reason that learning to output complex artifacts conditioned on observations of the environment would be a prerequisite to ongoing open-ended innovation. 

However, in preliminary experiments in the Sodarace domain, learning tended to ``gravitate'' towards collapsed solutions, wherein a single program is produced that achieves reasonable performance on a subset of the terrains in the distribution support. To reduce the viability of such an outcome and simulate a scenario where conditionality is essential, a small and discrete set of terrains for which a single program \emph{cannot} achieve good performance provides a test where conditional solutions should be significantly more advantageous.

\begin{figure}[t]
\centering
    \begin{subfigure}{.49\textwidth}
      \centering
      \includegraphics[width=\linewidth]{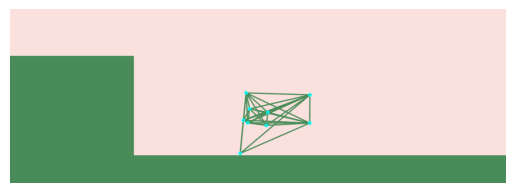}
      \caption{\textsc{left-wall}}
      \label{fig:left-wall}
    \end{subfigure}
    \hfil
    \begin{subfigure}{.49\textwidth}
      \centering
      \includegraphics[width=\linewidth]{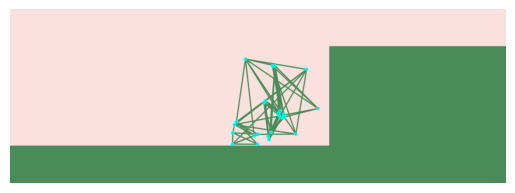}
      \caption{\textsc{right-wall}}
      \label{fig:right-wall}
    \end{subfigure}
    \\
    \begin{subfigure}{.49\textwidth}
      \centering
      \includegraphics[width=\linewidth]{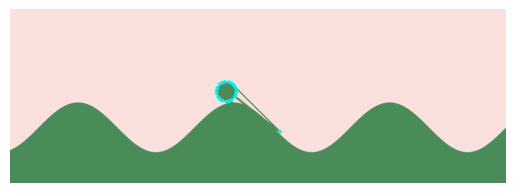}
      \caption{\textsc{bumpy}}
      \label{fig:bumpy}
    \end{subfigure}
    \hfil
    \begin{subfigure}{.49\textwidth}
      \centering
      \includegraphics[width=\linewidth]{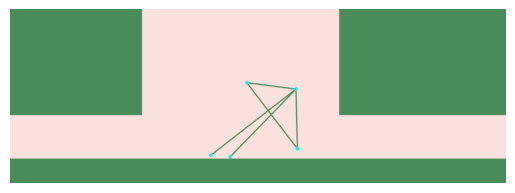}
      \caption{\textsc{tunnel}}
      \label{fig:tunnel}
    \end{subfigure}%
    \caption{\textbf{Terrains used in experiments}. A small set of terrains from which distributions that force \emph{conditionality} can be constructed. The terrains are (a) \textsc{left-wall}, (b) \textsc{right-wall}, (c) \textsc{bumpy}, and (d) \textsc{tunnel}. The Sodaracers produced by the models are incapable of scaling the walls in \textsc{left-wall} and \textsc{right-wall}, and therefore must produce different Sodaracers for these two terrains. Similarly, achieving good performance in the \textsc{tunnel} terrain can only be achieved with short Sodaracers, which struggle to locomote as quickly as taller ones, encouraging the model to distinguish between these terrains. Finally, Sodaracers that are proficient in locomotion on flat terrains tend to perform poorly on \textsc{bumpy}, encouraging the model to produce yet another Sodaracer for this terrain. In contrast to \textsc{tunnel}, which requires Sodaracers with a particular morphology, achieving good performance on \textsc{bumpy} requires modifying the way Sodaracers \emph{locomote}. Example Sodaracers are added to the figures to illustrate the scale of the terrains.}
    \label{fig:simple-terrain-distribution}
\end{figure}

In the experiments, uniform distributions are considered over sets of terrains as illustrated in \Cref{fig:simple-terrain-distribution}. Two subsets are considered, both of which contain \textsc{left-wall} and \textsc{right-wall}. One set additionally contains \textsc{tunnel}, and the other includes \textsc{bumpy}. These sets were specifically chosen such that the models are incapable of producing a single Sodaracer that achieves good performance on all terrains; to maximize the learning objective, the model must leverage the TEN to incorporate conditionality. %

\subsection{Parametrizing TENs}

\label{sec:ten-parametrization}
Two parametrizations for the TEN are explored. 
\paragraph{Discrete Codes.}
The terrain distribution has a discrete and finite support. 
As such, a simple parametrization wherein the terrains are treated as additional tokens in the existing vocabulary, and the embedding for each terrain is learned separately may be used.
The advantage of such a parametrization is that it introduces a relatively small number of new parameters to be optimized 
with RL, and it is conceptually simple to understand and debug.
However, the main disadvantages of such a parameterization are that
\begin{inlinelist}
    \item the number of parameters scales with the size of the terrain set, and 
    \item it does not allow the model to naturally generalize to unseen terrains at test-time, which may be an important constraint for downstream open-ended processes.
\end{inlinelist}

\paragraph{ResNets.}
An alternative parametrization is visual representations of the terrains, which can then be processed by visual recognition models. 
In particular, a ResNet50 \citep{he:arxiv15} embeds images into $\sR^d$ as a TEN when experimenting with visual representations of terrains.
The main advantages of this parametrization are that it is quite general, could conceivably be used in multiple settings (e.g. teaching a code-generating LLM to write programs in response to visual input, and in theory can generalize to unseen terrains.
The main drawback of this approach is that it introduces a large number of new parameters that must be optimized using a sparse RL signal.
Conversely, for large terrain distributions, this approach makes it possible to amortize 
the number of additional parameters necessary for designing conditional inventors.

\subsection{Experimental Details and Results}
\label{sec:conditional-rl-experiments}

Each RL episode consists of sampling a batch of terrains from the distribution, producing  samples %
from the conditional LLM, and evaluating them %
in simulation to produce the reward. 

\begin{figure}
\centering
    \begin{subfigure}{\textwidth}
      \centering
      \includegraphics[width=\linewidth]{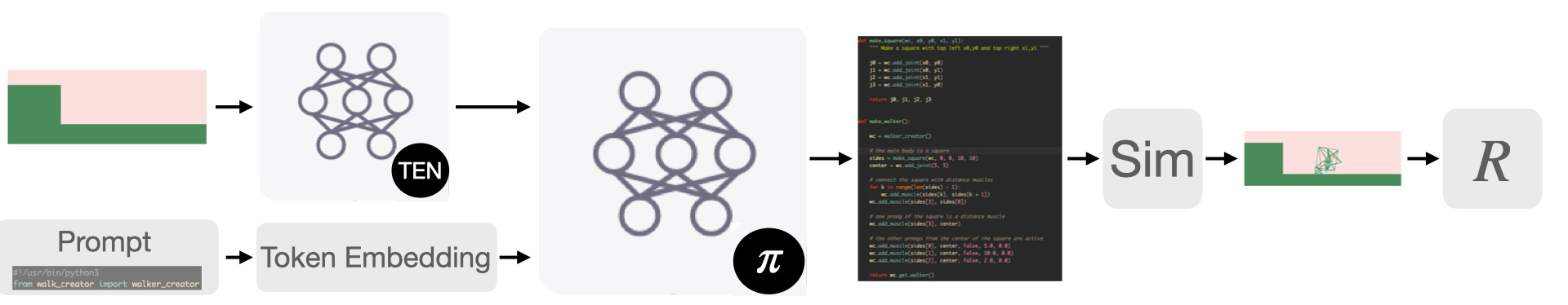}
      \caption{Policy network}
      \label{fig:policy-diagram}
    \end{subfigure}
    \par\bigskip
    \begin{subfigure}{\textwidth}
      \centering
      \includegraphics[width=\linewidth]{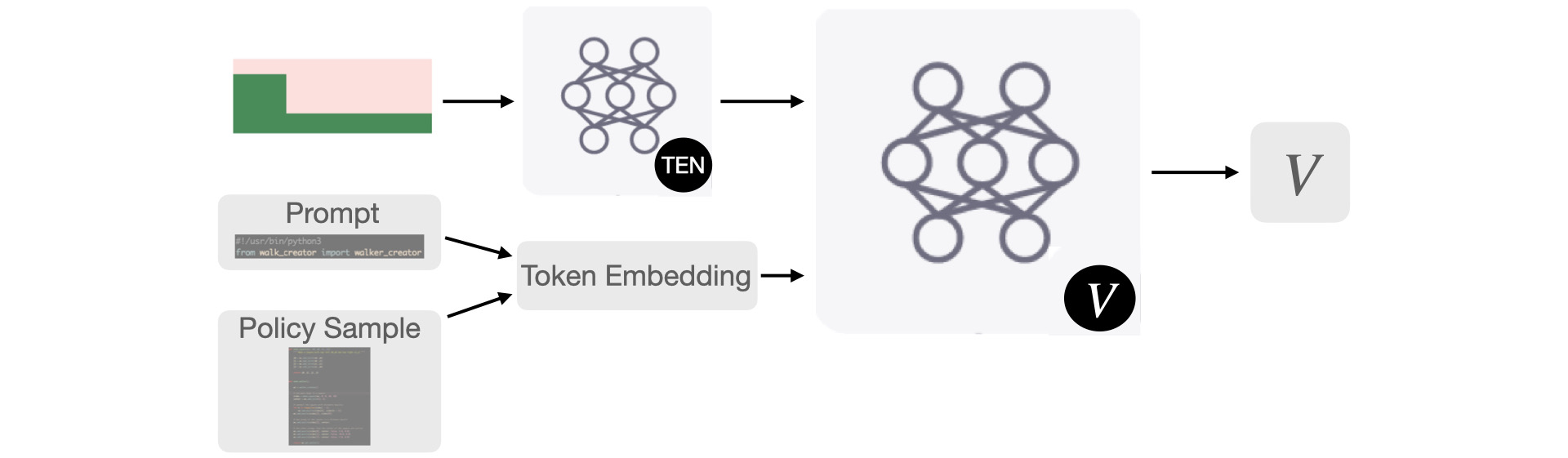}
      \caption{Value-function network}
      \label{fig:value-function-diagram}
    \end{subfigure}
    \caption{\textbf{Illustration of the RL architecture.} The conditional policy (a) and value-function (b) are depicted, both augmented with a separate TEN for terrain embeddings. The policy is conditioned on a particular terrain (via the TEN) and prompt, and produces a sample, which is interpreted as Python code. The code is then used to compile a Sodaracer, which is evaluated in a simulation to procuce a reward $R$. The value function is conditioned on the same terrain (via its own TEN) and prompt, and outputs an estimation of the value ($V$) of every token output by the policy sample. During learning, the value-function is trained to predict advantages estimated using GAE \citep{schulman2015high}.}
\label{fig:rl-architecture}
\end{figure}

Proximal policy optimization \citep[PPO;][]{schulman:ppo} is the RL algorithm, in conjunction with generalized advantage estimation \citep[GAE;][]{schulman2015high}, with default hyper-parameters.
In preliminary experiments, we found it important to add a KL term (between the policy network and the pre-trained LLM from Stage 2) \emph{to the reward function}, as proposed by \citet{christiano2017deep} and \citet{stiennon2020learning}. The value network is parametrized as a scalar-function version of the policy network, i.e.\ a separate LLM with a separate prepended TEN initialized from the Stage 2 models.
\Cref{fig:rl-architecture} illustrates the architectures and pipelines for the policy and value-function networks. Each iteration consists of batches of 1{,}024 samples (distributed over 32 GPUs), and training runs consist of 100 iterations. 

\begin{figure}
\centering
    \begin{subfigure}{.85\textwidth}
      \centering
      \includegraphics[width=\linewidth]{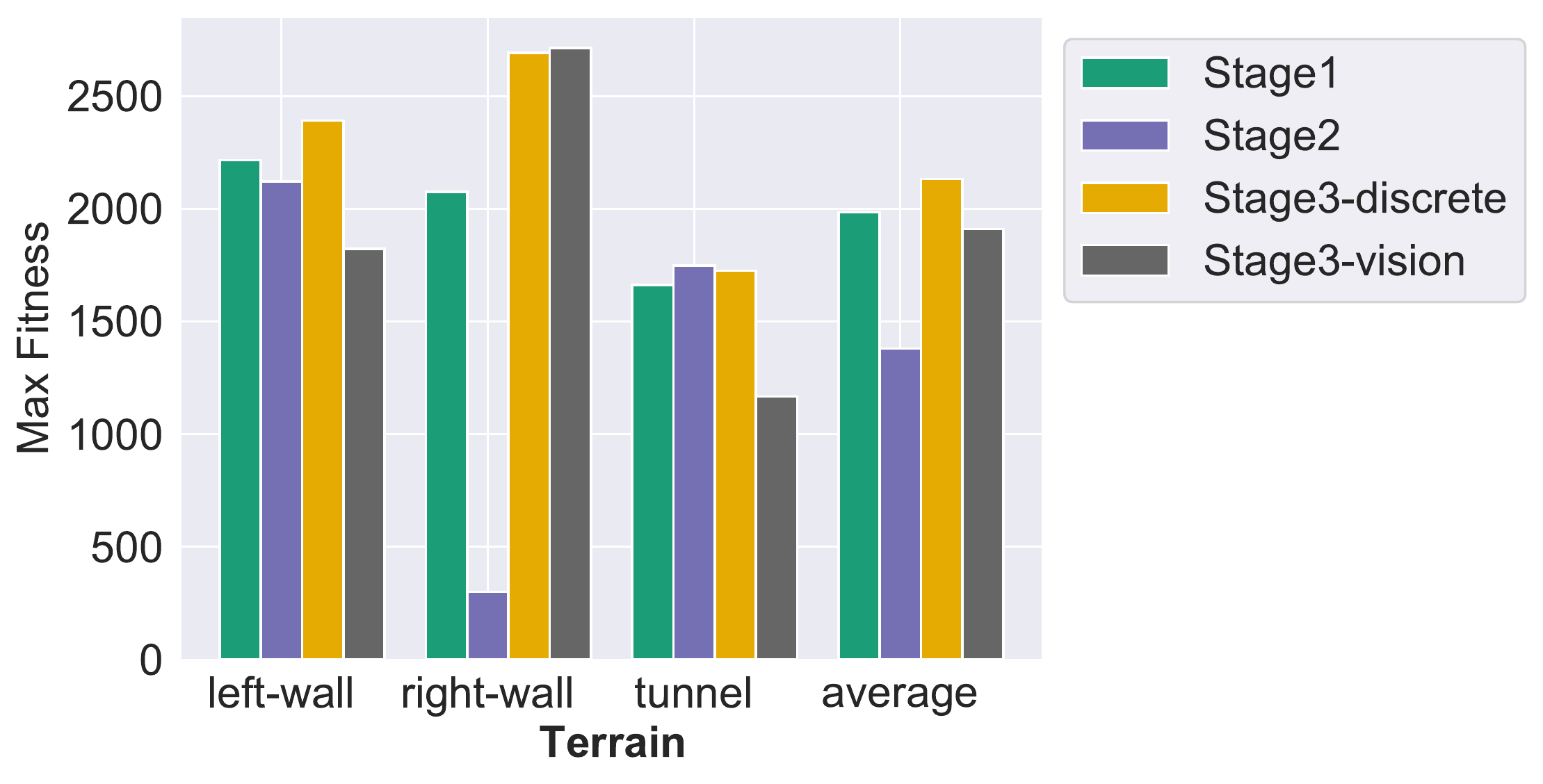}
      \caption{Cutoff 50\%}
      \label{fig:bumpy-perf-across-stages-50}
    \end{subfigure}
    \\
    \begin{subfigure}{.85\textwidth}
      \centering
      \includegraphics[width=\linewidth]{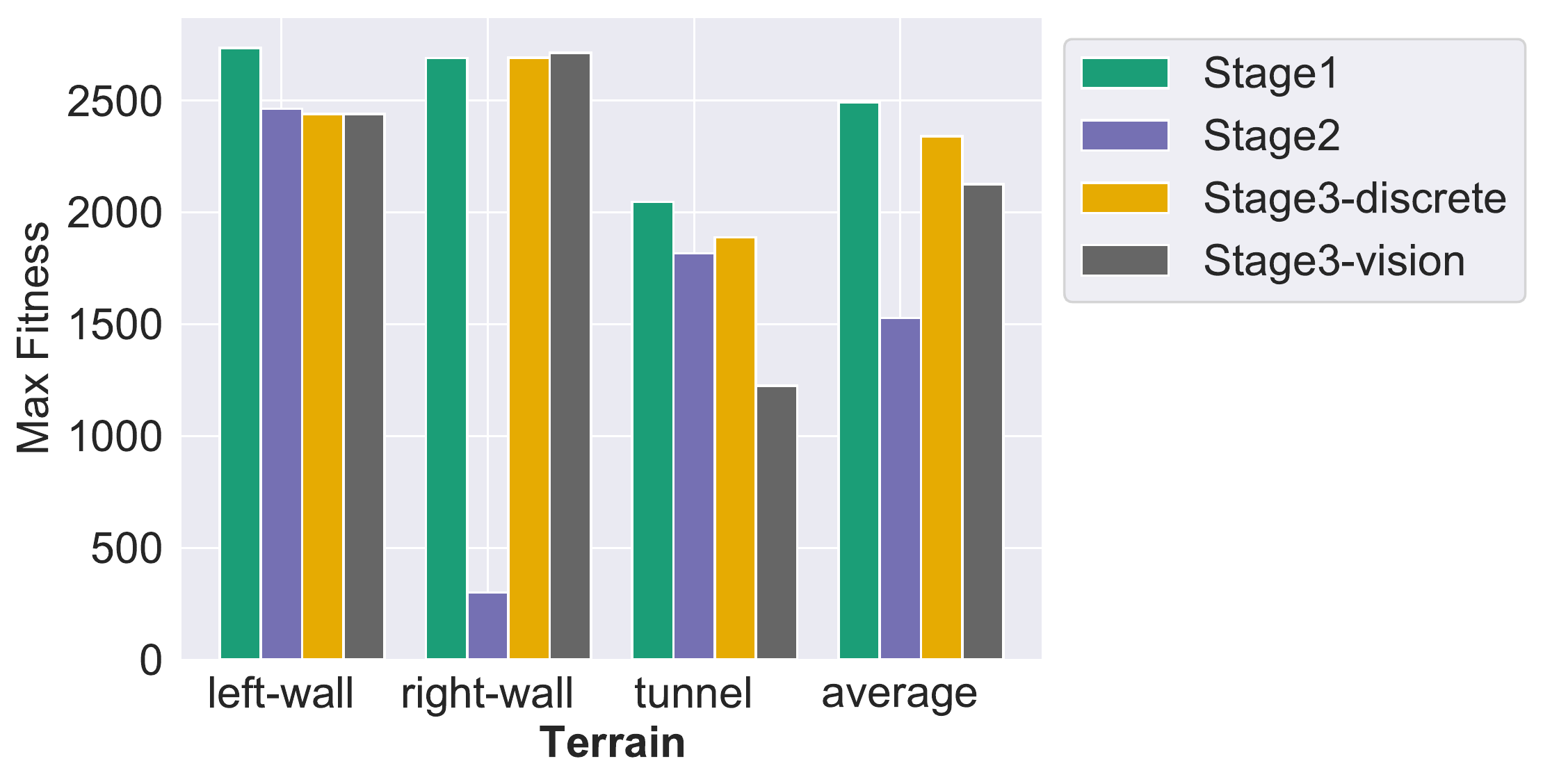}
      \caption{Cutoff 80\%}
      \label{fig:bumpy-perf-across-stages-80}
    \end{subfigure}
    
    \caption{\textbf{Comparing performance of models and datasets across the stages of the pipeline on the terrain distribution including the \textsc{tunnel}}. Results are detailed when training the LM using a dataset with a cutoff of (a) 50\%, and (b) 80\%. Stage 3 models are able to discover conditional solutions in both cases, consistently perform  comparably to test-time compute on the dataset, and better than the Stage 2 pretrained LMs. For the 80\% cutoff threshold, while performance is better than with 50\% at all stages, the pipeline struggles to improve performance over that of the dataset. Conversely, for the 50\% cutoff threshold, the Stage 3 (discrete) model improves upon all stages, demonstrating the ability of the pipeline to improve the performance of the models. 
    }
\label{fig:perf-across-stages-tunnel}
\end{figure}

\begin{figure}
\centering
    \begin{subfigure}{.85\textwidth}
      \centering
      \includegraphics[width=\linewidth]{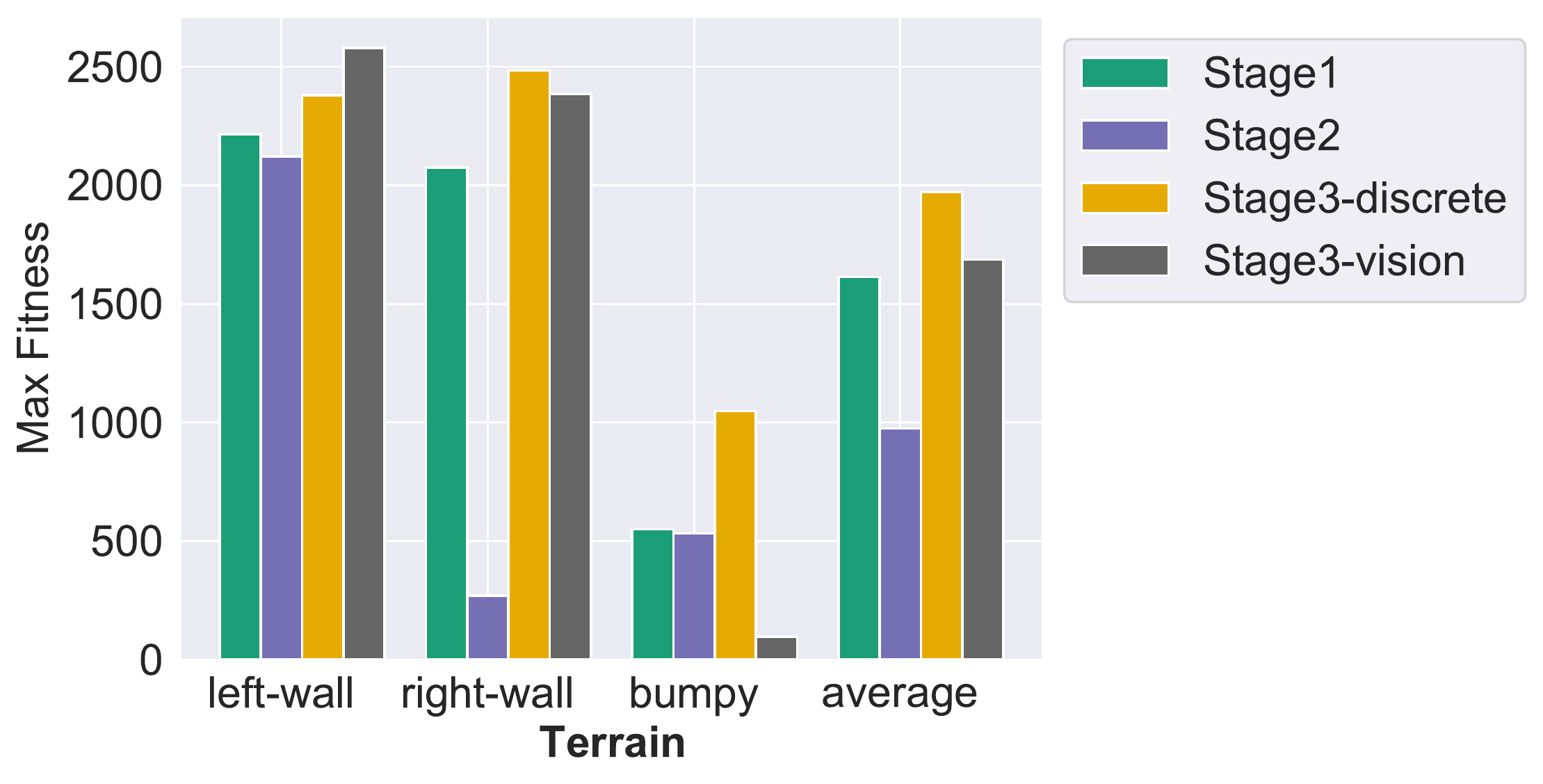}
      \caption{Cutoff 50\%}
      \label{fig:bumpy-perf-across-stages-50}
    \end{subfigure}
    \hfil
    \begin{subfigure}{.85\textwidth}
      \centering
      \includegraphics[width=\linewidth]{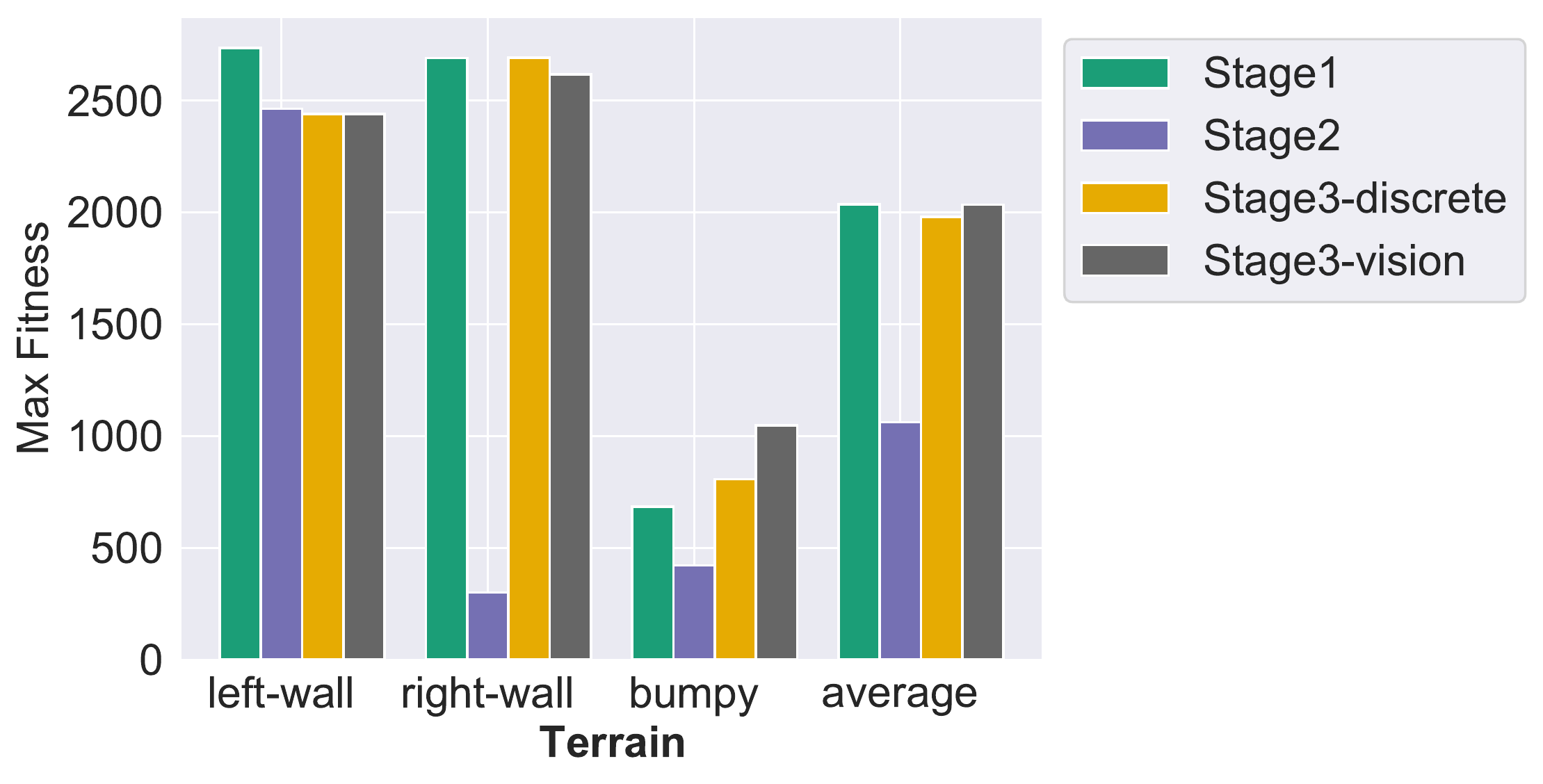}
      \caption{Cutoff 80\%}
      \label{fig:bumpy-perf-across-stages-80}
    \end{subfigure}
    
    \caption{\textbf{Comparing performance of models and datasets across the stages of the pipeline on the terrain distribution including \textsc{bumpy}.} Results are detailed when training the LM using a dataset with a cutoff of (a) 50\%, and (b) 80\%. Similar trends can be seen to those in \Cref{fig:perf-across-stages-tunnel}: Stage 3 models are able to discover conditional solutions and consistently perform comparably to test-time compute on the dataset, improving upon Stage 2 models. For the 80\% cutoff threshold, the pipeline achieves comparable performance to that of the dataset, while improving performance for the 50\% cutoff threshold.}
\label{fig:perf-across-stages-bumpy}
\end{figure}

RL is run on pretrained, 300M-parameter LLMs trained with datasets having cutoff thresholds in \{50\%, 80\%\}. Recall that we use the cutoff threshold to control the tradeoff between data quality and quantity, such that higher thresholds result in smaller pretraining datasets with a higher density of quality instances. For each dataset and terrain distribution combination, three runs are performed using different seeds, and the performance is averaged over samples from the resulting model for each terrain, from over all runs, though we exclude a small number of runs that diverged during training. To compute a measure of the performance of datasets and pretrained LLMs, we invoke test-time compute: 1{,}024 Sodaracers are sampled uniformly and evaluated from each dataset/model (recall that there is one model for both cutoff thresholds), and the best-performing Sodaracer is considered for each terrain. \Cref{fig:perf-across-stages-tunnel,fig:perf-across-stages-bumpy} detail the results of these experiments with the \textsc{tunnel} and \textsc{bumpy} distributions, respectively.

In short, \Cref{fig:perf-across-stages-tunnel,fig:perf-across-stages-bumpy} help us understand whether RL is able to discover conditional solutions, which we interpret as conditional inventors of Sodaracers that are capable of locomoting on particular terrains. Moreover, \Cref{fig:perf-across-stages-tunnel,fig:perf-across-stages-bumpy} enable us to compare the performance of Sodaracers produced at different stages of the pipeline, and how performance is affected by the choice of cutoff threshold. A particularly interesting question is whether RL is able to consistently improve upon the performance of test-time compute with the pretrained models produced in Stage 2. %

The RL procedure was at times brittle: training sometimes diverged, and some results were inconsistent. 
Divergence tended to be more frequent using the ResNet TENs, which is unsurprising considering the ResNets introduce many more parameters to the model, which are in turn trained with an extremely impoverished distribution of images (one for each terrain in the distribution).  %

Despite the fragility, RL fine-tuning is successful in producing conditional inventors in this domain: 
the models tend to produce a single Sodaracer for each terrain, which differ across terrains in the distribution. Importantly, the produced Sodaracers achieve good performance for the conditioned terrain, while failing to locomote on the other terrains. Videos showcase Sodaracers invented for the Tunnel distribution -- \url{https://y2u.be/e53NwdT4RdM} -- and for the bumpy distribution -- \url{https://y2u.be/WEM1dBtLLTw}. In short, the main result is the outputs of Stage 3, and thus the complete pipeline, are conditional inventors of the desired form.

Moreover, in most cases, the RL models are comparable to or better than the best-performing Sodaracers sampled from the dataset or the pretrained LLM. This consistency implies that Stage 3 enables the models to learn to use the TENs in conjunction with the LLMs, and further can fine-tune the models' outputs to improve performance, though not always 
by significant margins. %

Models trained with a cutoff of 80\% tend to achieve slightly better performance, and proved more stable during training, though the differences are not significant. This result implies that the tradeoff between data quality and quantity may play a role in downstream tasks (such as RL fine-tuning), a point that warrants further investigation in future work. One interesting avenue for research in this direction is to consider pretraining procedures that include information regarding the quality of the instances (where such information is available), e.g.~as proposed by \citet{chen2021decision}. 

Finally, we note that ``collapsed'' solutions in which the same Sodaracer is produced every time a particular terrain is observed (as opposed to significantly different samples each time the same terrain is seen) are sensible in this setting, as there should exist a dominant Sodaracer for each terrain. 
However, interestingly, in true open-ended systems this property may not hold: if the environment is constantly shifting, that excludes the existence of single, dominant inventions. In such a setting, the stochasticity of the model is expected to be beneficial, enabling the model to adapt and produce a diversity of useful solutions.

\subsection{Qualitative Observations}
\label{sec:wheel}

Several interesting structures and solution classes were qualitatively observed throughout the experiments, which provide additional insight into the pipeline’s ability to conditionally invent solutions to different terrains. One such example is the emergence of very \emph{short} Sodaracers, which arose in response to the \textsc{tunnel} terrain. The video visualizations at \url{https://y2u.be/P9A1ruI3_tU}  highlight examples of such Sodaracers produced in response to \textsc{tunnel}.

Another interesting class of Sodaracers appeared in earlier experiments with ELM; a wheel-like structure emerged during the evolutionary process, and persevered throughout the pipeline. During Stage 3, the wheel proved particularly adept at locomoting in the \textsc{bumpy} terrain, and consistently emerged as the solution to \textsc{bumpy} produced by the Stage 3 models for that terrain across RL runs. Unfortunately, the wheel did not re-emerge in the ELM runs used in the final experiments in this paper. The video at \url{https://y2u.be/l5PVSLDknWM}  demonstrates several solutions of this form discovered by RL when trained with the \textsc{bumpy} terrain distribution as well as the \textsc{tunnel} distribution. For contrast, this video (\url{https://y2u.be/Mo-rXnFq6vQ})  show failure modes on bumpy for several Sodaracers effective in locomoting on flat terrains.

Such qualitative observations provide further evidence that the pipeline is capable of producing interesting inventors and creative solutions to problems, even in a simplified domain that is not open-ended. We hypothesize that when unleashed in more complex domains, this capability of conditional invention will contribute to the open-endedness of the induced process by continually introducing new objects to the environment, and thus changing its properties for other agents.

\section{Discussion and Conclusion}

An important difference between natural evolution and most of EC is the very beginning--nature began with a single ``example'' or seed, the first cell on Earth, that was already bestowed with critical initial functionality and information.  In contrast, runs in EC usually begin with randomized configurations with little or no useful information. 
Because programming languages like Python for humans are natural modalities for formalizing complicated ideas and relationships, such a program could serve as a seed more in the spirit of nature.  However, the problem then is that arbitrary mutations to an already-formulated program are very unlikely to be useful.  

A few years ago, the idea that the mutation operator could ``know'' how to perturb such programs in reasonable and promising ways would be fanciful, but, as shown in this paper. the emergence of LLMs has now made such capabilities a reality. The MAP-Elites algorithm combined with ELM easily bootstraps datasets of hundreds of thousands of examples in a completely foreign domain (to the initial LLM) from initial human-written seeds.  The validity of this generated data is confirmed by the invention pipeline that follows--conditional LLMs were ultimately trained starting from this data that cannot be trained from scratch.  

More broadly, the main idea introduced here is that LLMs trained on code open up a significant new kind of intelligent GP enabled by ELM that is no longer at the mercy of the raw search landscape induced by code.  While the experiment in this paper points to a set of implications for open-endedness, deep learning, and RL, the potential applications are numerous and many previous challenges in the GP field could be revisited with this new tool.     

The experiment in this paper shows that intelligent LLM-based mutation operators 
can successfully drive exploration by being combined with other search algorithms (e.g.\ MAP-Elites in this work). Furthermore, optimizing such mutation operators based on the quality of their output during the search itself appears to make them work even better for exploration.
Not only are the discoveries of such search potentially useful in their own right (like wheels in the Sodarace domain), but they offer an entirely new option for generating example data or optimizing existing solutions in domains where data is sparse or non-existent. For example, such search through LLM-based perturbation could feasibly be applied to optimize the MAP-Elites search algorithm itself, or for LLM architecture and hyperparameter search.

From the perspective of open-endedness, the challenge in principle is that the search is by definition continually and even intentionally shifting out of distribution.  As soon as a new invention or DCT is achieved, open-endedness demands that its now-familiar comfort zone be at least partially abandoned for new frontiers. The experiment here wherein LLMs trained from simple flat-ground walkers were able to leverage that knowledge to appropriately generate specialized walkers for different terrains shows just this kind of informed leap to a new frontier.  If such a process of leaps upon leaps can be made to continue indefinitely, then an unbounded explosion of emergent complexity could be within reach.

One important question for future work is the extent to which the resultant model can interpolate or extrapolate to examples (i.e.\ environments) outside its training distribution. 
While RL can harness existing knowledge in the LLM to bootstrap into new tasks, extrapolating principles from such knowledge is much harder and likely to require further weight updates through additional learning.  It is possible that a sophisticated future open-ended system would entangle both continual evolution and RL for DCTs together.

Overall, the hope is that the simple core insight that the efficacy of mutation in GP can now dramatically improve through ELM will inspire a broad array of novel applications and research directions.  The observation that EC can benefit directly and dramatically from advances in deep learning (and deep learning from EC further down the invention pipeline) can also help to motivate the further pursuit of synergies between the fields.

\section*{Acknowledgments}

Thank you to Jeff Clune for substantive insightful feedback and thoughtful discussion about the project and this paper. Thanks also to Glenn Powell for consistent useful input and ideas during team meetings and discussions. We also thank the Supercomputing team for their work, which enabled our experimentation.

\bibliography{big,new,cites,nn,oee1,oee2,ucf, gordonjo}

\appendix
\section{Comparing Mutation Operators}
\label{appendix:intelligent_perturbation}

This section gives more details on experiments using different mutation operators to fix bugs in a single step of perturbation. A simple form of GP mutation is implemented with Tiny GP \cite{Sipper2019tinyGP}, a tree-based GP implementation. In particular, the mutation operator is restricted to mutating nodes, which offers it the highest chance of success given the nature of the bugs introduced (which do not introduce new structure, but instead each bug in effect swaps an incorrect node for a correct one). For Tiny GP, The mutation rate was tuned by hand for each problem. While many more sophisticated mutation operators in GP exist \cite{salustowicz1997probabilistic,spector2002genetic}, the motivation for these experiments is mainly to highlight the potential for LLM-based directed mutation operators to make sophisticated movements along the manifold of code. 

The experimental setup for each task (described next) is that each mutation operator is given many independent trials, where the operator perturbs a single buggy parent (with potentially several bugs), and the resulting child is tested for correctness. First, a Python 3 version of each function was written, then (for perturbation by GP) it was translated by hand into a Tiny GP tree. The commit message for the diff models for these tasks is ``Fixed bugs.'' Note that GP cannot make use of the plain-text doc-string that describes what the function is intended to do, which highlights another advantage of LLMs for perturbation, in that they can use (and create) language-based comments to guide the evolution of code. For prompt engineering, the following prompt format was used (with ``\{problem\}'' replaced with the buggy implementation code):

\begin{minted}[mathescape,
               linenos,
               numbersep=5pt,
               gobble=0,
               frame=lines,
               framesep=2mm]{python}
# A buggy implementation
{problem}

# Fixed Bugs
def 
\end{minted}

Two benchmark tasks were explored: 4-Parity, where the objective is to calculate the parity of a 4-bit sequence, and Quadratic, where the objective is to calculate the result of a quadratic function, given the values of the coefficients $a$, $b$, $c$, and the independent variable $x$. The motivation for 4-Parity is that bit parity is a 
common GP benchmark task \cite{koza:book92} and provides a simple test-bed for whether LLMs can make multiple coordinated (and effective) changes to code. Quadratic provides another simple test-bed, and unlike 4-Parity, the bugs introduced are more ambiguous (i.e.\ the function description does not imply that the function must be of any canonical form), making it more similar to the use case of this paper, wherein undirected yet semantically meaningful changes are desired. Note that the nature of the introduced bugs for each task are described in the next section along with the source code for the correct implementation (into which bugs are introduced).

\subsection{Python Source for Benchmark Tasks}

\subsubsection{4-Parity}

\begin{minted}[mathescape,
               linenos,
               numbersep=5pt,
               gobble=0,
               frame=lines,
               framesep=2mm]{python}
#!/usr/bin/python3
def parity(b1,b2,b3,b4):
    """ Return binary parity of a sequence of input bits. 
        Return 0 for even parity, 1 for odd parity """
    bit_sum = sum([b1,b2,b3,b4])
    return bit_sum %
\end{minted}

Bugs were incrementally introduced in the following way: For the first four mutations, each variable with a ``b`` prefix was renamed with a ``c`` prefix (e.g. first ``b1'' is changed to ``c1'', then ``b2'' is changed to ``c2'', etc.). For the fifth mutation, the ``modulus two'' is replaced with modulus three. For GP, additional ``c''-prefixed terminals were introduced. 

\subsubsection{Quadratic}

\begin{minted}[mathescape,
               linenos,
               numbersep=5pt,
               gobble=0,
               frame=lines,
               framesep=2mm]{python}
#!/usr/bin/python3
def quadratic(a,b,c,x):
    """ Return quadratic: a,b,c are coefficients 
        and x is the independent variable."""
    return a*pow(x,2)+b*x+c
\end{minted}

A maximum of two bugs was introduced to the Quadratic task by individually replacing the $+$ operators with $-$ operators, from left to right. 

\subsection{Comparing GP and Diff Mutation}

\begin{figure}[t]
\centering
\includegraphics[width=0.8\textwidth]{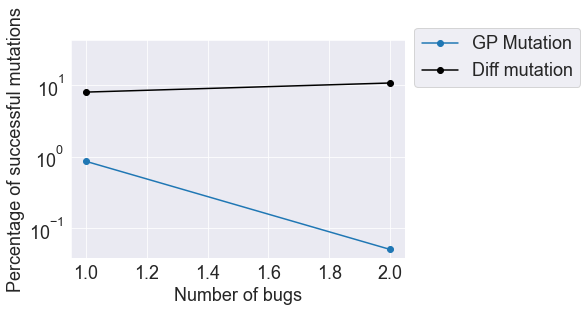}
    \caption{\textbf{Comparing diff mutation to GP mutation at fixing bugs in the Quadratic task.} 
    The plot shows how the ability of a single mutation to produce correct solutions changes as bugs are incrementally
    added to a working implementation that solves the Quadratic task. Note that success percentage is shown in \emph{log scale}, i.e.\ success for
    GP mutation decreases significantly when the second bug is added, while diff mutation is unaffected. The conclusion is that this task adds more evidence that LLM-based mutation can make
    multiple sensible coupled changes to code.
    }
    \label{fig:intelligent_mutation_quadratic}
\end{figure}

Performance plots that compare GP to diff mutation for the 4-Parity task are in Figure \ref{fig:intelligent_mutation} in the main text. Figure \ref{fig:intelligent_mutation_quadratic} in this appendix shows the same comparison for the Quadratic task, where performance is similar to 4-Parity, i.e.\ the diff mutation's performance is both greater than GP mutation for both cases of bugs, and degrades in a different pattern (here, the performance of diff mutation is unaffected by the introduction of the second bug).

\subsection{Comparing API-based Mutations} \label{sec:appendix-api-comps}

\begin{figure}[t]
\centering
\includegraphics[width=0.8\textwidth]{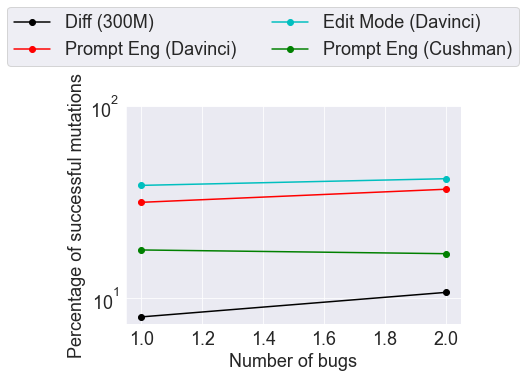}
    \caption{\textbf{Comparing alternate LLM-based mutations at fixing bugs in the Quadratic task.} 
    The performance of different mutation operators in fixing bugs is shown as bugs are incrementally added to an correct implementation for the Quadratic task. Both edit mode and prompt-engineering approaches outperform the 300M diff model applied in this paper's experiments. The conclusion is that this additional task adds evidence that there exist multiple viable options to build upon the work in this paper.
    }
    \label{fig:api_quadratic}
\end{figure}

Performance plots that compare diff mutation to mutations possible through the OpenAI API for the 4-Parity task can be seen in Figure \ref{fig:api_parity} in the main text. Figure \ref{fig:api_quadratic} here shows the same comparison for the Quadratic task, which highlights similar results: There are multiple options for mutation operators available through the OpenAI API that perform as well or better than the diff model applied in this paper's experiments.

\section{Seed Source Code}
\label{appendix:seed_source_code}

This section contains the source code of the seed programs for ELM. Videos for the Sodaracers these programs represent are available at: \url{https://y2u.be/jeP8Nsulu48}.

\subsection{CPPN Seeds}

There are two CPPN-like seeds. CPPN-Fixed does not allow the core functionality of the CPPN encoding (encapsulated in the \textrm{query\_cppn} function) to
change, whereas CPPN-Mutable includes the source code for that function, thereby enabling the CPPN encoding itself also to evolve.
\label{source:cppn}

\subsubsection{CPPN-Fixed}

\begin{minted}[mathescape,
               linenos,
               numbersep=5pt,
               gobble=0,
               frame=lines,
               framesep=2mm]{python}
def make_walker():
    wc = walker_creator()

    def connect(x1,y1,x2,y2):
        if ((x1-x2)**2+(y1-y2)**2)>4.5:
            return False
        return True

    def amp(x1,y1,x2,y2):
        return max(abs(x1-x2),abs(y1-y2))

    def phase(x1,y1,x2,y2):
        return np.sign(x1)

    joints = query_cppn(wc,8,3,1.5,connect,amp,phase)

    return wc.get_walker()
\end{minted}

\subsubsection{CPPN-Mutable}

\begin{minted}[mathescape,
               linenos,
               numbersep=5pt,
               gobble=0,
               frame=lines,
               framesep=2mm]{python}
def query_cppn(wc, xgrid,ygrid,scale,connect_func,amp_func,
               phase_func):
    """ Create a grid of points and functionally connect them. """
    joints = {}
    for x in range(xgrid):
        for y in range(ygrid):
            joints[(x,y)] = wc.add_joint(x*scale,y*scale)

    for x1 in range(xgrid):
        for y1 in range(ygrid):
            for x2 in range(x1,xgrid):
                for y2 in range(y1,ygrid):
                    if x1==y1 and x2==y2:
                        continue
                    if connect_func(x1,y1,x2,y2):
                        amp = amp_func(x1,y1,x2,y2)
                        phase = phase_func(x1,y1,x2,y2)
                        wc.add_muscle(joints[(x1,y1)],
                            joints[(x2,y2)],False,amp,phase)

    return joints

def make_walker():
    wc = walker_creator()

    def connect(x1,y1,x2,y2):
        if ((x1-x2)**2+(y1-y2)**2)>4.5:
            return False
        return True

    def amp(x1,y1,x2,y2):
        return max(abs(x1-x2),abs(y1-y2))

    def phase(x1,y1,x2,y2):
        return x1 if x1%

    joints = query_cppn(wc,8,3,1.5,connect,amp,phase)
\end{minted}

\subsection{Square Seed}
\label{source:square}

\begin{minted}[mathescape,
               linenos,
               numbersep=5pt,
               gobble=0,
               frame=lines,
               framesep=2mm]{python}
def make_square(wc, x0, y0, x1, y1):
    """ Make a square with top left x0,y0 and top right x1,y1 """

    j0 = wc.add_joint(x0, y0)
    j1 = wc.add_joint(x0, y1)
    j2 = wc.add_joint(x1, y1)
    j3 = wc.add_joint(x1, y0)

    return j0, j1, j2, j3


def make_walker():

    wc = walker_creator()

    # the main body is a square
    sides = make_square(wc, 0, 0, 10, 10)
    center = wc.add_joint(5, 5)

    # connect the square with distance muscles
    for k in range(len(sides)-1):
        wc.add_muscle(sides[k], sides[k+1])
    wc.add_muscle(sides[3], sides[0])

    # one prong of the square is a distance muscle
    wc.add_muscle(sides[3], center)

    # the other prongs from the center of the square are active
    wc.add_muscle(sides[0], center, False, 5.0, 0.0)
    wc.add_muscle(sides[1], center, False, 10.0, 0.0)
    wc.add_muscle(sides[2], center, False, 2.0, 0.0)

    return wc.get_walker()
\end{minted}

\subsection{Radial Seed}
\label{source:radial}

\begin{minted}[mathescape,
               linenos,
               numbersep=5pt,
               gobble=0,
               frame=lines,
               framesep=2mm]{python}
def make_circle(wc, cx,cy,radius,num_points):
    """ Approximate a circle with center (cx,cy) square with 
        num_points points """
    joints = []

    tot_ang = 3.14*2.0

    for idx in range(num_points):
        ang = tot_ang/(num_points-1)*idx
        x = math.cos(ang) * radius + cx
        y = math.sin(ang) * radius + cy
        joints.append(wc.add_joint(x,y))

    return joints


def make_walker():

    wc = walker_creator()

    num_points = 8
    rad = 5.0
    cx,cy = (5,5)
    # the main body is a square
    points = make_circle(wc, cx,cy,rad,num_points)
    center = wc.add_joint(cx,cy)

    for k in range(num_points):
        wc.add_muscle(points[k], points[(k+1)%
        wc.add_muscle(points[k], center,False,float(k)/num_points,
                      float(k)/num_points)

    return wc.get_walker()
\end{minted}

\section{Model Architectures}
\label{appendix:arch}

Architectural details for the language models applied in this paper are shown in Table \ref{table:arch}. Models are based on the GPT-3 architecture, and further description of
architecture and hyperparameters can be found in \citet{brown2020language}.

\begin{table}[ht!]
\centering
\begin{tabular}{|c c c c c|} 
 \hline
 $n_\textrm{params}$ & $n_\textrm{layers}$ & $d_\textrm{model}$ & $n_\textrm{heads}$ & $d_\textrm{head} $ \\ [0.5ex] 
 \hline\hline
 0.1M & 2 & 64 & 4 & 16 \\ 
 85M & 12 & 768 & 12 & 64 \\
 350M & 24 & 1{,}024 & 16 & 64 \\
 760M & 24 & 1{,}536 & 16 & 96 \\
 \hline
\end{tabular}
	\caption{\textbf{Model architectures.} The table shows hyperparameters that describe the architectures of the models used in this paper, including parameter count ($n_\textrm{params}$), number of layers ($n_\textrm{layers}$), number of units in each bottleneck layer ($d_\textrm{model}$), number of attention heads ($n_\textrm{heads}$), and dimension of each attention head ($d_\textrm{head}$).}
\label{table:arch}
\end{table}

\section{Seed Robustness}
\label{appendix:robustness}

A subtle issue came to light when bringing together the full pipeline, which is that
there are complex interactions between the kind of seed that kicks off ELM
in Stage 1 and the performance of RL models trained in Stage 3.  In particular,
some seeds (like the Radial seed) attain high QD scores in Stage 1, but fail to provide good jumping-off
points to adapt to novel terrains in Stage 3. When examining the products of the Radial seed, 
many of them exhibited chaotic dynamics that appeared overly sensitive to initial conditions. Similarly chaotic
results were observed with the CPPN-Mutable seed trained with the pretrained diff model. The conclusion
is that QD score does not entirely capture what enables generalization and
 adaptation to novel terrains. Understanding this issue may be important for further research.

Possible ideas for biasing seeds towards producing generalizable
inventions include disallowing precise setting of joint position and
oscillatory parameters, introducing stochasticity to prevent overfitting to initial
conditions, and incrementally adjusting the seed.
Preliminary results in disallowing precise setting of parameters provided mixed results.

One promising result came from incremental seed design.
With the CPPN-Mutable seed (where the logic describing the CPPN encoding was able to
be evolved), the pretrained diff model behaves similarly to the Radial
seed (it creates inventions with high quantitative performance but which exploit 
chaotic dynamics). However, when the diff model is fine-tuned on the products of the
CPPN-Fixed seed (where the core CPPN logic is conserved), further
CPPN-Mutable runs retain qualitative characteristics of the CPPN-Fixed
seed while outperforming it quantitatively. That is, the CPPN-Fixed seed provided
``training wheels'' for learning how to modulate the encoding itself in the CPPN-Mutable seed.
In this way, an incremental
approach to seed design (potentially involving interactive evolution) may be a promising approach to qualitatively
shaping the outputs of ELM; alternatively, the notion of QD score
could be expanded or changed to better align with robust downstream performance.

\section{Final Map Approach to Stage 2}
\label{appendix:finalmap}

There are a variety of ways to distill the raw data generated by Stage 1 into
a dataset upon which a model can be trained. This section details a natural
alternative approach to the percentage threshold method used in the paper, called the final 
map approach. The method is to concatenate from all runs the solutions from
their final MAP-Elites maps, i.e.\ the best quality solutions for each niche
at the end of a run. 

This approach strikes a different trade-off between quantity and quality of
data samples than the percentage threshold method. The percentage threshold
approach normalizes performance across runs for each niche, and then includes
all reasonably-high quality solutions. The final map approach, on the other hand,
is agnostic to the performance of a
given run or seed (it does not normalize across runs), and for each run takes only the highest-quality data
for each discovered niche.
 
The final map dataset naturally consists of fewer examples (only 13K examples).
Models trained on the final map generally perform worse than
percentage threshold models on QD score. Lower QD results from the fact that
the performance across the final map varies significantly across seeds (e.g.\ the Square seed performs very strongly
in certain niches, but fails to find solutions in others, while the CPPN-like seed discovers solutions in nearly all
niches, but generally with weaker performance). As a result, the average sample from the final map dataset performs
worse than those from the percentage threshold dataset (resulting in lower QD score in the dataset, and also in trained
models).

Additionally, preliminary Stage 3 experiments
proved unstable when using models trained on the final map dataset. In effect, the final map
dataset appears to be too small to serve as a reliable jumping-off point for
further RL. 

\section{Source Code for Completion Targets}
\label{appendix:completion_targets}

This section includes the source code for the three inventions that serve as out-of-distribution completion tests for
trained models in Stage 2. Videos for these inventions is shown at: \url{https://y2u.be/8C2K5fk28HI}.

ELM often
adds structure to the seed, as in the nested loop of the Wheel, or the multiple added loops in the Galloper, and also
reuses function calls (e.g. calling \textrm{make\_sensor} several times in the Galloper; note that \textrm{make\_sensor} is a renamed (and modified) version
of the \textrm{make\_square} function included in the Square seed.

Nonsensical comments are often inserted (as in ``acrylic of current (m)" in the Runner's source), although
 parsimony pressure in the MAP-Elites algorithm tends to eventually
strip them out (e.g.\ there are no comments in the Wheel invention). In some situations the seed's original comments
are preserved, as in the comment ``connect the square with distance muscles" in the source code of the Galloper. 

\subsection{Wheel}

\begin{minted}[mathescape,
               linenos,
               numbersep=5pt,
               gobble=0,
               frame=lines,
               framesep=2mm]{python}
import math
def make_circle(wc, cx,cy,radius,num_points):
    joints = []
    tot_ang = 3.14*2.0
    for idx in range(num_points):
        ang = tot_ang/(num_points+1) * idx
        x = math.cos(ang) * radius + 0.5
        y = math.sin(ang) * radius + cy
        joints.append(wc.add_joint(x,y))
    return joints 
def make_walker():
    wc = walker_creator()
    num_points = 8
    rad = 3.0
    cx,cy = (11,5)
    points = make_circle(wc, 0.6, -0.5,rad/2,num_points)
    center = wc.add_joint(cx+1,cy+1)
    for j in range(num_points):
        for i in range(num_points-5):
            wc.add_muscle(points[j], points[(i+j)%
                          0.0, 1.0, (j+1)/num_points)
        wc.add_muscle(points[j], center,False,3,(j+1)/num_points)
    return wc.get_walker()
\end{minted}

\subsection{Galloper}
\begin{minted}[mathescape,
               linenos,
               numbersep=5pt,
               gobble=0,
               frame=lines,
               framesep=2mm]{python}
def make_sensor(wc, x0, y0, x1, y1, d):
    return wc.add_joint(x0, y0), wc.add_joint(x1, y1), 
           wc.add_joint(x1, y0), wc.add_joint(x0, y1), 
           wc.add_joint(d, 0.5), wc.add_joint(x1, 0.5)
           
def make_walker(dx=0.0, dy=0.0, ddr=0, ddc=1.6, sid=8.0,
                s_influence=0.2, s_side_width=0.0, 
                first_center=5.0, last_center=15.0):
    wc = walker_creator() 
    ends = [make_sensor(wc, 5 + dx, -1 + dy, ddr, ddc, 4.5),
        make_sensor(wc, 0, -0.1, sid, 9.5, 0.03),
        make_sensor(wc, 5.5, -0.001, 5.0, 4.86 +0.8, 0.07),
        make_sensor(wc, 5.5, -3.0, 6.0, 4.86 + 0.8, 0.07),
        make_sensor(wc, 0, dx, ddr, ddc, 1.0)]

    sides = ends[0] + ends[1] + ends[2] + ends[-1] + ends[-2] 
            + ends[-3]

    center = wc.add_joint(dx, dy)
    # connect the square with distance muscles
    for k in range(len(sides)-6):
        wc.add_muscle(sides[k], sides[k+1], True, 30, 0.5)
    wc.add_muscle(sides[2], sides[4], False, 4.0, 0.8)
    for k in range(len(sides)-2):
        wc.add_muscle(sides[k], sides[k + 2], True, 18.0, 
                      60.0 / 5.5)

    for k in reversed(range(len(sides)-6)):
        wc.add_muscle(sides[k], sides[k + 5], False, 4.0, 
                      20.0 / 9.0)
    wc.add_muscle(center, sides[7], False, 2.0, 90.0 / 9.0)
    return wc.get_walker()
\end{minted}

\subsection{Runner}

\begin{minted}[mathescape,
               linenos,
               numbersep=5pt,
               gobble=0,
               frame=lines,
               framesep=2mm]{python}
import math
import numpy as np

def make_walker(p_scale=1):         # acrylic of current (m)
    wc = walker_creator()

    def connect(x1,y1,x2,y2):
        if -2*x1+x2*2>2:
            return True
        return x1<= abs(y1-y2)

    def amp(x,y,x2,y2):
        return abs(x-x2) + abs(y-y2)

    def phase(x1,y1,x2,y2):
        return -x1/2 - math.cos(math.pi/9)

    joints = query_cppn(wc,5,7+p_scale,2,connect,amp,phase)
    return wc.get_walker()
\end{minted}

\section{Source Code for Selected Stage 1 Sodaracers}
\label{appendix:source_code}

\subsection{Blob (from CPPN Seed)}

A video of the Sodaracer represented by the code below can be seen at: \url{https://y2u.be/JDUAI8yrNcY}.

\begin{minted}[mathescape,
               linenos,
               numbersep=5pt,
               gobble=0,
               frame=lines,
               framesep=2mm]{python}
import math

def walker():
    wc = walker_creator()

    def connect(x1,y1,x2,y2):
        return (x1-x2)**2+5*y1**2-4*x2**2+y2**2 > 2.5
    def amp(x1,y1,x2,y2):
        return (x1-x2)**2+x2**2 + 1 - y2**2 < 2

    def phase(x1,y1,x2,y2):
        return math.sin(x1)*math.cos(y1)**2 + 1

    joints = query_cppn(wc,5,6,2.1,connect,amp,phase)
    return wc.get_walker()
\end{minted}

\subsection{Hopper (from Square Seed)}

A video of the Sodaracer represented by the code below can be seen at: \url{https://y2u.be/noSPGFX5m3M}.

\begin{minted}[mathescape,
               linenos,
               numbersep=5pt,
               gobble=0,
               frame=lines,
               framesep=2mm]{python}
def make_square(wc, x0, y0, x1, y1, length):
    j0 = wc.add_joint(x0, y0)
    j1 = wc.add_joint(x0, y1)
    j2 = wc.add_joint(x1, y1)
    j3 = wc.add_joint(x1, y0)

    return j0, j1, j2, j3


def make_walk(n=6):

    wc = walker_creator()

    # the main body is a square
    sides_2_theta = make_square(wc, 0.0, 0.0, 5.6, 9.4, 2.4)
    sides_1_theta = make_square(wc, 0.5, 0.8, 6.5, 13.1, 1.3)
    sides_2_theta += make_square(wc, -0.8, -0.6, 6.7, 13.0, 2.3)
    sides_2_theta += make_square(wc, -0.9, -0.6, 8.4, 12.5, 0.7)
    sides_2_theta += make_square(wc, 0.0, -0.5, 0.2, 12.4, 1.7)
    sides = sides_1_theta + sides_2_theta + sides_1_theta
    center = wc.add_joint(2, 2)

    # connect the square with distance muscles
    for k in range(len(sides)-2):
        wc.add_muscle(sides[k], sides[k+1])
        wc.add_muscle(sides[k+2], sides[k], False, 30.0, 30.0)

    # similarities of the Squares with":
    for k in range(len(sides)-2):
        wc.add_muscle(sides[k], sides[k], True)

        for n in range(k, len(sides)):
            wc.add_muscle(sides[k], sides[n], False)
    wc.add_muscle(sides[3], center)
    # the other prongs from the center of the square are active
    wc.add_muscle(sides[2], center, False, 25.0, 25.0-0.7)
    wc.add_muscle(sides[3], center, False, 20.0, 30.0+0.4)

    return wc.get_walker()
\end{minted}

\subsection{Centipede (from Radial Seed)}

A video of the Sodaracer represented by the code below can be seen at: \url{https://y2u.be/zhMsPzo22do}.

\begin{minted}[mathescape,
               linenos,
               numbersep=5pt,
               gobble=0,
               frame=lines,
               framesep=2mm]{python}
import math


def make_circle(wc, cx,cy,radius,num_points,eccentricity=1.4):
    joints = []


    tot_ang = math.pi*2.0*eccentricity

    for idx in range(1,num_points):
        x = math.cos(3.14*(idx+num_points)*tot_ang/(num_points)) 
                     * radius + cx
        y = math.sin(3.14*(idx+num_points)*tot_ang/(num_points)) 
                     * radius + cy
        joints.append(wc.add_joint(x,y))

    return joints

def make_walker(num_points=300,rad=3.25,f=3,max_rad=3):
    wc = walker_creator()

    cx,cy = (0,0)
    body_size = rad*1.625

    points = make_circle(wc, 0,0,body_size,num_points)
    center = wc.add_joint(cx,cy)

    for k in range(1,num_points-1):
        wc.add_muscle(points[((k%
                      int(f*k/float(10)), k/10.)
        wc.add_muscle(points[(k%

    return wc.get_walker()
\end{minted}

\section{Probing Stage 2 Models}
\label{appendix:probing}

One hope for the models trained in Stage 2 is that they will learn not
only to memorize the training data (Python examples of Sodaracers), but also to internalize the underlying structure of the domain (e.g. how in general to mix together springs and masses to create functional Sodarace inventions). 
This section discusses some preliminary observations of informal experiments that change the training procedure in Stage 2 to explore what the model is capable of learning. In particular,
Sodarace examples are augmented with additional comments (either as a prefix or postfix) that contain both the Sodaracer's fitness and its behavior characterization (its width, height, and mass).

The idea is that after training, the model can be asked to
predict e.g.\ the fitness of an unseen invention (if trained with postfix comments), or to generate a walker with desired properties (if trained with prefix comments).
For example, a prefix-trained model can be conditionally sampled based on a prefix that specifies the desired height, width, and mass of a Sodaracer, to see how reliably samples can match those properties when evaluated in the domain.

Preliminary experiments with both prefix and postfix 300M parameter models highlighted that the model was able to make such associations within the training distribution, e.g.\ when a postfix model was queried with heights, widths, and masses taken from test set examples (held out from the same distribution), it was able to consistently generate a Sodaracer with those properties. It was slightly less reliable when conditioned on fitness, reflecting that this is a much more complicated association (e.g.\ unlike width and height, fitness depends on the physical dynamics of the generated walker).

However, when taken out of distribution the model
was less robust. For example, a prefix model struggled to targetedly generate Sodaracers within a band of width and height that was deliberately held out from the training set. Interestingly, while it was not reliable in generating Sodaracers of particular held-out widths and heights, samples from the model did in effect cover
the holdout area, suggesting that the variation accessible within the model is
enough for interpolation or slight extrapolation, which is an important property
for enabling continual open-ended elaboration.

More starkly, a postfix model had very limited ability to predict the
fitness of Sodaracers taken from the Radial seed, which was not seen in training (there was a Spearman correlation of only $0.08$). One hypothesis that is left
to future work to explore, is that larger
models, trained with much more generated data, may have more robust performance
when taken out-of-distribution. If true, this would support that scaling can benefit open-ended learning, just as it does in unsupervised and supervised learning.

A more speculative line of thought emerging from these experiments relates to
how Stage 2 \emph{structures} the knowledge about the domain, which may significantly
impact the dynamics of how RL in Stage 3 unfolds. That is, by training the model
to associate Sodaracers with their properties (through a prefix or postfix), it may
be more likely that Stage 3 can smoothly interpolate in the space of those properties, which otherwise the model would have no explicit knowledge about. However,
when a prefix-trained model was tested in the interpolation setup of 
Appendix \ref{appendix:interpolation}, it did not perform any better than those trained without prefixes. While such prefix-training did not have the desired impact, it remains
an open question how to include within Stage 2 information that intuitively seems highly-relevant to RL (like fitness) in a way that maximally benefits such RL.

Overall, the conclusion is that (at least with 300M parameter models and the current amount of training data), Stage 2 models demonstrate modest capabilities to learn structure within Sodarace, but are not yet robust when taken out-of-distribution. The implication for open-endedness is unclear (whether or not this poses a problem for future research): For example, it may be that stronger generalization capabilities may more naturally emerge when the existing pipeline (which is mainly a proof of concept) is extended such that Stage 3 is embedded within an open-ended process. 
Indeed, at least in human processes of innovation, general
insight does seemingly often emerge from continual open-ended 
accumulation of initially-disparate examples of phenomena that are only later unified.

\section{Interpolation Experiments}
\label{appendix:interpolation}

This section discusses experiments probing how well the conditional inventor (the product of Stage 3) is able to understand the domain of the invention, by exploring whether the model can appropriately adjust its output in response to structured changes in the environment. That is, adjusting inventions in response to smooth variations in the environment requires a deeper understanding of the structure of the domain, and could potentially enable inventors to generalize beyond the environments observed during training. 

To examine this capability, an environment distribution with smoothly varying features is created, in particular, by varying the heights of \emph{tunnel} terrains. 
The motivation for this distribution is the observation that while larger Sodaracers are unable to navigate low tunnels, they tend to locomote more quickly on flat terrains. Thus, the model is incentivized to adapt the height of the produced Sodaracer to the height of the tunnel in the terrain, using ``taller" Sodaracers that locomote quickly for the taller tunnels, and shorter, slower-moving Sodaracers for the lower tunnels. The ability to achieve such a solution would imply that the model has learned about the underlying structure of the domain, in that it is able to tweak the height of the produced inventions, and has captured this relationship between the height and speed of the Sodaracer. To enable the model to %
potentially learn a smooth mapping from the height of the tunnel to the produced Sodaracer, the ResNet TEN architecture is employed.

In the experiments, however, the model repeatedly converged on solutions outputting the same Sodaracer regardless of the height of the tunnel, i.e.\ an unconditional solution. Examples of such solutions are shown at \url{https://y2u.be/gt1Z0lnjAuE}.%

These results point towards a subtle characteristic of the invention pipeline introduced in this work. The models are not exhibiting a deep understanding of the domain, finding a local, unconditional optimum that works ``reasonably" well on almost all terrains in the distribution. Particularly concerning is that the produced Sodaracer is not able to navigate all terrains in the distribution, highlighting the suboptimality of the learned solution. This property confounds probing the interpolation capabilities of the inventors, and it remains unclear if the invention pipeline can produce complex solutions that are able to smoothly vary the produced inventions in response to smooth variations in the environment. Conversely, the experiments presented in the main body of this document imply that the model is able to produce conditional solutions when no unconditional solutions are sufficient.

We speculate that unconditional local optima are simpler and easier to learn using RL methods, such that the models ``gravitate" towards them when such solutions exist. However, in future work, the invention pipeline could be deployed in more complex, open-ended processes where unconditional solutions should be rendered insufficient. In such settings, it is conceivable that the pipeline will output conditional inventors that have a deeper understanding of the domain structure, as such solutions will allow the inventors to achieve significantly higher rewards in the domain, negating the concern regarding unconditional solutions. 

Another avenue for future research would attempt to make the learning task posed in Stage 3 easier by exploring maximum likelihood learning methods when bootstrapping the conditional inventor (Stages 1 and 2). Here, the assumption is that the exploration task in Stage 3, coupled with the necessity of incorporating the new modality, is quite challenging for RL procedures. A simple approach to this could be to sample from the unconditional LLM multiple times, and use the best-performing samples for each terrain in the distribution as a supervised (terrain-Sodaracer pairs) dataset to fine-tune both the LLM and the TENs. Stage 3 could include terrain-distributions incorporating terrains unseen during Stage 2, encouraging the inventor to further generalize and explore the space of inventions. Looking even further down the line, it is conceivable to replace the MAP-Elites-based ELM procedure of Stage 1 with a POET-style algorithm \cite{wang2020enhanced}, which would produce a supervised dataset of this form during Stage 1, relieving pipeline designers of the need to hand-specify terrain distributions on which to train the conditional inventor in Stage 2.

\end{document}